\documentclass[10pt, twocolumn, journal,web]{article}
\usepackage{array}
\usepackage[caption=false,font=normalsize,labelfont=sf,textfont=sf]{subfig}
\usepackage{textcomp}
\usepackage{stfloats}
\usepackage{url}
\usepackage{verbatim}
\usepackage{graphicx}
\usepackage{cite}
\usepackage{pifont}
\usepackage[colorlinks,
            linkcolor=blue,
            anchorcolor=blue,
            citecolor=blue]{hyperref}

\usepackage{float} 
\usepackage[caption=false]{subfig}

\usepackage[linesnumbered,ruled,vlined]{algorithm2e} 
\usepackage{setspace}

\usepackage{booktabs}
\usepackage{multirow}
\usepackage{bbding}

\usepackage{hyperref}
\usepackage{amsmath,amssymb,amsfonts}

\usepackage{cleveref}

\usepackage{tabularx}
\usepackage{times}
\usepackage[margin=16mm]{geometry}
\usepackage{amsmath,amssymb,amsfonts}
\usepackage{subfig}
\usepackage[utf8]{inputenc}
\usepackage{utfsym}
\usepackage{multirow}
\usepackage[normalem]{ulem}
\usepackage{amsmath,amssymb,amsfonts}
\usepackage{algorithmic}
\useunder{\uline}{\ul}{}
\usepackage{url}
\def\BibTeX{{\rm B\kern-.05em{\sc i\kern-.025em b}\kern-.08em
    T\kern-.1667em\lower.7ex\hbox{E}\kern-.125emX}}
\usepackage{amssymb}
\usepackage{cite}
\usepackage{graphicx}
\usepackage{authblk}

\makeatletter
\renewcommand\AB@affilsepx{, \protect\Affilfont}
\makeatother
\usepackage{tabularx}
\providecommand{\keywords}[1]
{
  \small	
  \textbf{\textit{Keywords---}} #1
}

\providecommand{\keywords}[1]{%
  \small
  \textbf{\textit{Keywords---}} #1%
}

\begin{document}

\title{\textbf{Local and Global Context-and-Object-part-Aware Superpixel-based Data Augmentation for Deep Visual Recognition}}

\author[1, 2,  *]{Fadi Dornaika}
\author[1]{Danyang Sun}
\affil[1]{\textit{University of the Basque Country}}
\affil[2]{\textit{IKERBASQUE}}
\affil[ ]{

\small\texttt{fadi.dornaika@ehu.eus, danyangsun@163.com}}
\date{}

\maketitle

\begin{abstract}
Cutmix-based data augmentation, which uses a cut-and-paste strategy, has shown remarkable generalization capabilities in deep learning. However, existing methods primarily consider global semantics with image-level constraints, which excessively reduces attention to the discriminative local context of the class and leads to a performance improvement bottleneck. Moreover, existing methods for generating augmented samples usually involve cutting and pasting rectangular or square regions, resulting in a loss of object part information. To mitigate the problem of inconsistency between the augmented image and the generated mixed label, existing methods usually require double forward propagation or rely on an external pre-trained network for object centering, which is inefficient. To overcome the above limitations, we propose LGCOAMix, an efficient context-aware and object-part-aware superpixel-based grid blending method for data augmentation. To the best of our knowledge, this is the first time that a label mixing strategy using a superpixel attention approach has been proposed for cutmix-based data augmentation. It is the first instance of learning local features from discriminative superpixel-wise regions and cross-image superpixel contrasts. Extensive experiments on various benchmark datasets show that LGCOAMix outperforms state-of-the-art cutmix-based data augmentation methods on classification tasks, {and weakly supervised object location on CUB200-2011.} We have demonstrated the effectiveness of LGCOAMix not only for CNN networks, but also for Transformer networks. Source codes are available at \url{https://github.com/DanielaPlusPlus/LGCOAMix}.
\end{abstract}

 \hspace{10pt}

\section{Introduction}
Deep learning has fostered tremendous advances in image classification \cite{chan2015pcanet,zhang2023spatial}, object detection \cite{kong2020foveabox,zou2023object}, and image segmentation \cite{chen2017deeplab,minaee2021image} due to its ability to extract effective information from large external data sources. And as the amount of data increases, deep learning is further promoted, especially for Vision Transformers (ViTs) \cite{dosovitskiyimage,khan2022transformers}. However, annotating data is time-consuming, costly, or infeasible. This is the main reason why we have enormous amounts of data, but not enough data for deep learning. 
\begin{figure}[ht!]
	\centering
	\includegraphics[scale=0.55]{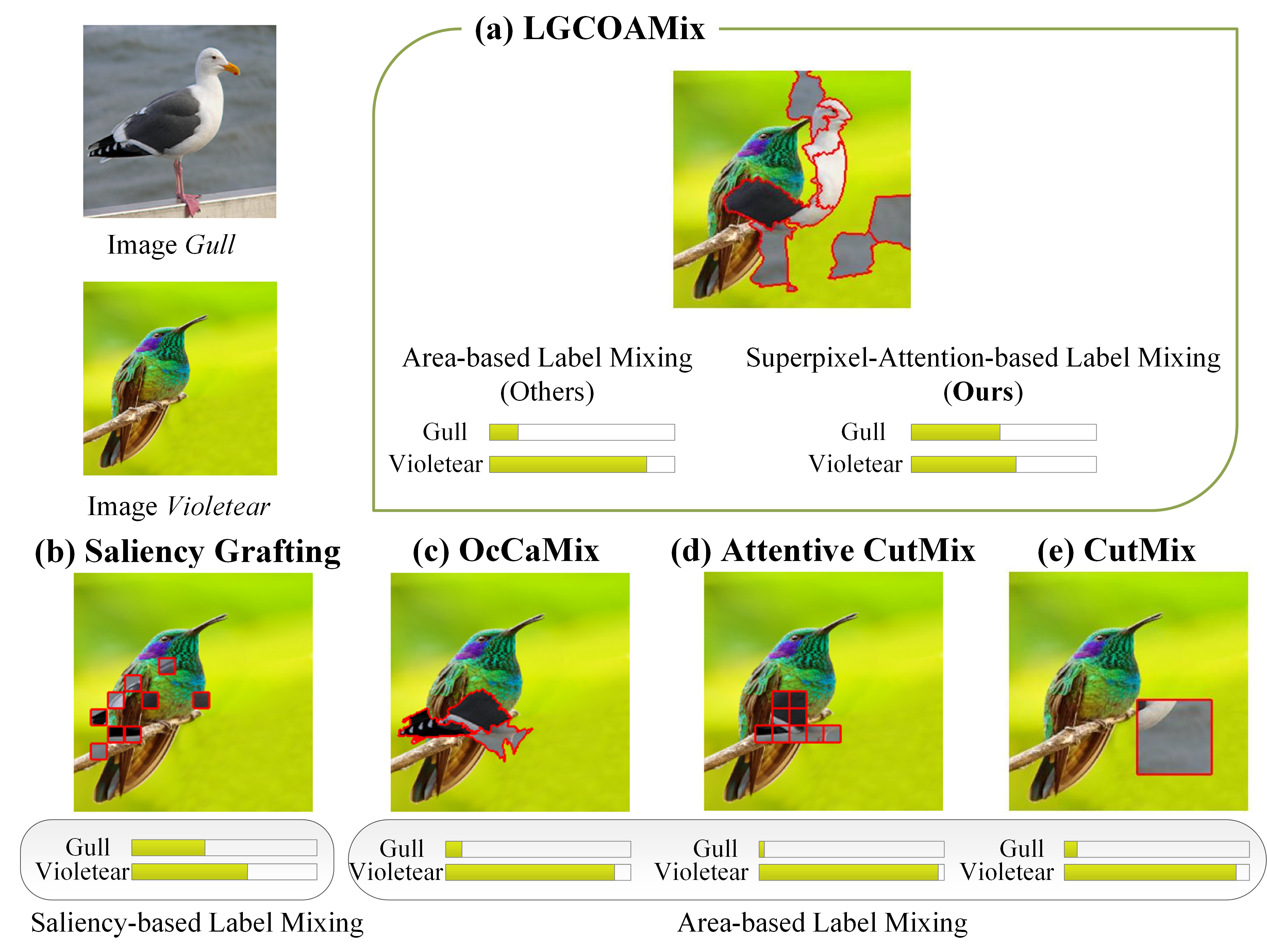}
    \caption{Comparison of augmented samples and label mixing methods.
 (a) LGCOAMix generates local object-part preserving augmented samples with superpixel-attention-based label mixing with a single forward propagation, which is more semantic and efficient than area-based label mixing. (b) \cite{park2022saliency} uses saliency-based label mixing, but local object part information is lost because the mixing is in square form. (c) \cite{dornaika2023object} and (d) \cite{walawalkar2020attentive} use area-based label mixing with object centering. (c) \cite{dornaika2023object} requires double forward propagation. (d) \cite{walawalkar2020attentive} requires an external pre-trained network. In (d), the local object part information is also lost. (e) \cite{yun2019cutmix} encounters inconsistencies between the augmented image and the generated mixed label and loses the local object part information.}
    \label{fig1}
\end{figure}

To alleviate the problem of data scarcity, data augmentation \cite{xu2023comprehensive} is the most well-known solution. In recent years, one of the most popular data augmentation ideas is cutmix-based with a cut-and-paste strategy. Cutmix-based data augmentation has been shown to improve generalization and localization by allowing training samples to represent not only the most discriminative regions but the entire object\cite{yun2019cutmix}. However, existing cutmix-based data augmentation methods have three drawbacks: 
 (\uppercase\expandafter{\romannumeral1}) Existing methods successfully extend the focus from discriminative regions to the entire image. However, they only consider the global semantics with image-level constraints and overly reduce the focus on the discriminative local context, resulting in a performance improvement bottleneck. (\uppercase\expandafter{\romannumeral2}) 
Existing methods typically produce augmented samples by mixing rectangular or square regions, resulting in a loss of object-part information in the augmented images. (\uppercase\expandafter{\romannumeral3}) Existing methods confront a dilemma between the need for diversification and concentration. Diversification is an essential requirement for data augmentation. Random mixing can contribute to diversification, but random mixing with background information and not class-related information leads to inconsistency between the augmented image and the label.
This is especially problematic when using the area-based label mixing method (Elaborated on in Sec. \ref{background}). To avoid this, some methods generate augmented samples by mixing discriminative regions with object centering \cite{walawalkar2020attentive,  liu2022automix,dornaika2023object,kim2020puzzle}. However, the deterministic concentration of discriminative regions leads to a lack of diversification. Moreover, many existing methods need to perform forward propagation twice\cite{liu2022automix,dornaika2023object,kim2020puzzle} or require an external pre-trained network for object centering\cite{walawalkar2020attentive}, which is inefficient.

To overcome the above limitations, we propose \textbf{LGCOAMix}, an efficient \textbf{L}ocal and \textbf{G}lobal \textbf{C}ontext-and-\textbf{O}bject-part-\textbf{A}ware superpixel-based grid mixing data augmentation with cut-and-paste strategy and a training framework for Deep Visual Recognition. The motivation is to improve deep encoder learning through image data augmentation.

By addressing the above shortcomings, our method is context-aware, object-part-aware, and efficient. First, we create augmented images by cutting and pasting local regions based on superpixels to obtain information about the object parts. Moreover, we mix the labels for the augmented images with the online superpixel-wise attention. Our proposed scheme mitigates the inconsistency problem between the augmented image and the label by adopting a real superpixel-based grid mixing and attention blending coefficients for the generated label. In this way, we only need a single forward propagation and get the maximum diversification, which is efficient. At the same time, we use superpixel pooling and self-attention to capture the contextual relationships and select the most relevant superpixels during training. Superpixel pooling preserves object-part information in the feature space. Based on the selected superpixels, local classification is then performed to further capture the local contextual information, and cross-image contrastive superpixel learning is used to achieve the alignment and consistency of the discriminative superpixel context.

Fig. \ref{fig1} shows a visual comparison between our method and other representative methods. Our method in Fig. \ref{fig1}(a) can generate the augmented samples with maximum diversification and preserve object-part local information. The superpixel-attention-based label mixing can perform label mixing semantically and accurately with an efficient single forward propagation.

Our main contributions are fourfold:

\begin{itemize}
\item  We discuss the potential shortcomings of existing cutmix-based data augmentation methods for image classification.

\item We propose an efficient object-part-aware superpixel-based grid mixing method for data augmentation. Unlike existing cutmix-based data augmentation methods, we propose for the first time a superpixel-attention-based semantic label mixing strategy that efficiently requires only a single forward propagation, does not require pre-trained modules, and performs label mixing semantically and accurately based on attention without destroying the augmentation diversification.

\item We propose a novel framework for training a strong classifier that is context and object oriented as well as efficient. To the best of our knowledge, this is the first instance of learning local features from discriminative superpixel regions and cross-image local superpixel contrasts.

\item We present extensive evaluations of LGCOAMix on several benchmarks and backbone encoders. These evaluations show that LGCOAMix outperforms existing cutmix-based methods for data augmentation.

\end{itemize}

\section{Related Work}
\label{related}

\textbf{Cutmix-based Data Augmentation.} 
Dropout \cite{srivastava2014dropout} inspired a training strategy by randomly deactivating the nodes of the fully connected networks for better generalization. In contrast to the original dropout regularization that operates on the model, regional dropout data augmentation operates on the image in the input space or feature space. Cutout \cite{devries2017improved} and Random Erasing \cite{zhong2020random} remove random regions of the image in the input space, DropBlock \cite{ghiasi2018dropblock} removes random regions of the image in the feature space. While they diversify the focus of the model, they also suffer from the loss of information that comes from removing regions directly.  Mixup \cite{zhang2018mixup} and Manifold Mixup \cite{verma2019manifold} randomly mix two images pixel by pixel for augmented images separately in input and hidden space. PuzzleMix \cite{kim2020puzzle} suggests a mixup method guided by saliency information. AutoMix \cite{liu2022automix} proposes a mixup framework with a parametric mix block and the momentum pipeline for training two subtasks. However, the mixup strategy, which mixes pixel by pixel, is difficult to interpret. CutMix \cite{yun2019cutmix} mitigates the information loss problem of Cutout \cite{devries2017improved} by cutting random regions from one image and pasting them into the other and derives the cutmix-based data augmentation. ResizeMix \cite{qin2020resizemix}, GridMix \cite{baek2021gridmix}, Random SuperpixelGridMix \cite{hammoudi2022superpixelgridmasks}, and PatchUp \cite{faramarzi2022patchup} generate the augmented images or features with complete randomness following CutMix \cite{yun2019cutmix}. However, the cut-and-paste strategy with complete randomness and area-based label mixing leads to the problem of inconsistency between the augmented image and the generated mixed label. Attentive CutMix \cite{walawalkar2020attentive}, PuzzleMix \cite{kim2020puzzle}, OcCaMix \cite{dornaika2023object} and AutoMix \cite{liu2022automix} propose to solve the inconsistency between the augmented image and the generated mixed label by attention or saliency guided region selection. However, they require forward propagation twice or additional networks, which is inefficient. Mixing with square regions causes the loss of object-part information. Saliency Grafting \cite{park2022saliency} generates the mixed label with semantic saliency-based label mixing method, but selects the regions in square form, which loses the information of the object part. Our approach, LGCOAMix, cuts and pastes the random superpixel regions to create an augmented image that preserves the object part information. To solve the problem of inconsistency between the augmented image and the generated mixed label, we propose a semantic superpixel attention-based label mixing strategy to create labels for the augmented images. Our proposed data augmentation method requires only a single forward propagation and is object-part aware.

\textbf{Pooling in CNNs.} 
Pooling in CNNs can downscale feature maps to achieve a larger receptive field and lower memory requirements. Common max pooling and average pooling layers have led to excellent performance in many classical convolutional models.
Recent work on pooling has focused on developing new pooling layers to better scale down feature maps. Learned-norm pooling \cite{gulcehre2014learned} uses a learned parameter $p$ to determine the relative importance for weighted sampling of activations within a kernel region. Spatial pyramid pooling \cite{he2015spatial} pools features in arbitrary regions (subimages) to create fixed-length representations for training, eliminating the need for fixed-size input images. S3pool \cite{zhai2017s3pool} uses the grid-sampling method, randomly sampling the rows and columns of the original feature map grid. Local importance pooling \cite{gao2019lip} uses learned weights as an attention-based mechanism for subnetworks to reweight patches for pooling. LiftPool \cite{zhaoliftpool} uses four different learnable sub-bands to generate the mixture of the sub-bands. Most of the above methods perform pooling based on square regions or kernels, resulting in loss of object part information. Our work proposes superpixel pooling and self-attention layer. Our pooling method can not only keep the object part information, but also weight the local context with the neighborhood relation.

\section{Proposed Approach}
\label{ours}

\begin{figure*}[h]{
	\centering
	\includegraphics[scale=0.38]{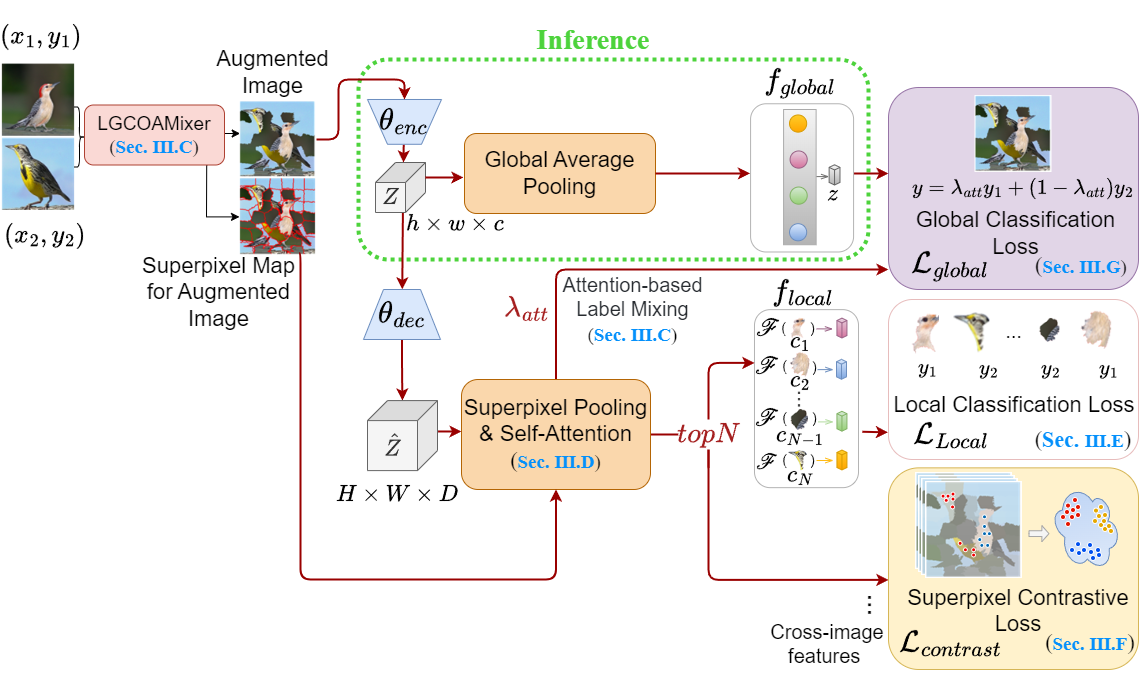}
    \caption{The overall framework of our LGCOAMix method.
    }\label{fig2}}
\end{figure*}

\subsection{Overview}
The overall framework of our LGCOAMix method is shown in Fig.~\ref{fig2}.
Two training images and their corresponding labels input LGCOAMixer (Detailed in Sec. \ref{ccamixer}). LGCOAMixer outputs the augmented image and its corresponding superpixel map. Then, the augmented image is fed into the encoder $\theta_{enc}$ and the decoder $\theta_{dec}$ for feature extraction and spatial resolution increase. 
The encoded low-resolution feature $\mathbf{Z}$ is used for global classification (Detailed in Sec. \ref{train-and-inference}) with semantic superpixel-attention-based label mixing (Detailed in Sec. \ref{ccamixer}) to train a strong classifier. The decoded high-resolution feature $\mathbf{\hat{Z}}$ is fed into the superpixel pooling and self-attention layer (Detailed in Sec. \ref{sap}) with the generated superpixel map to create a sequence of attentional superpixel vectors. The selected vectors with the top attention weights for discriminative superpixels are then used for local superpixel classification (Detailed in Sec. \ref{local-mapping}), as well as contrastive learning (Detailed in Sec. \ref{scl}). Only the well-trained encoder  $\theta_{enc}$ and the global classifier $f_{global}$ are used for inference.
We emphasize that our framework employs attentional mechanisms at three levels in the proposed architecture: (\romannumeral1) Attention-based label mixing for the global classification loss; (\romannumeral2) Self-attention for superpixel contextual learning; and (\romannumeral3) Attention-based superpixel selection for local superpixel classification and contrasts.

\subsection{Background}
\label{background}

Following CutMix \cite{yun2019cutmix}, traditional cutmix-based data augmentation methods typically cut regions from one image and paste them into another image to augment the images. For the labels of the augmented images, conventional methods mix the two labels with a certain proportion, which is called label mixing.

Let $x\in\mathbb{R}^{W\times{H}\times{C}}$ denote an arbitrary training image and $y$ is its corresponding class label. $W$ is the width, $H$ is the height, and $C$ is the number of channels. Cutmix-based data augmentation creates new augmented training image $(x_{mix},y_{mix})$  from two different training images $(x_1,y_1)$, $(x_2,y_2)$. 

The generation of the augmented images is described below.

\begin{equation}\label{eq1}
\centering
    x_{mix} = (\mathbf{1}-\mathbf{M})\odot{x_1} + \mathbf{M}\odot{x_2}
\end{equation}

$\mathbf{M}\in\{0,1\}^{W\times{H}}$ is a mixing binary mask of which pixels are part of $x_2$, $\mathbf{1}$ is the mask with value ones, $\odot$ means element-wise multiplication.

Label mixing for the augmented sample can be described as follows.

\begin{equation}\label{eq2}
\centering
    y_{mix} = (1-\lambda)\, {y_1} + \lambda \, {y_2}
\end{equation}

In the traditional cutmix-based method, $\lambda$ denotes the ratio  of the number of pixels cut from image $x_2$ and pasted into image $x_1$ to the total number of pixels in image $x_1$, which is area-based 
and described as Eq. (\ref{eq3}).

\begin{equation}\label{eq3}
\centering
    \lambda_{area} = \frac{\sum_{i=1}^{W}\sum_{j=1}^{H}{M}_{i,j}}{W\times{H}}
\end{equation}

In traditional cutmix paradigms that use areas to compute the coefficient $\lambda$ in Eq. (\ref{eq3}), the generated mixed label is assumed to be bimodal. This label may not be accurate because a source image can contribute with background information rather than class-specific information. And forward propagation is always required twice when we try to solve the inconsistency problems between the augmented image and the generated mixed label by image concentration, once for {object centering} and once for encoder training. Worse, deterministic object centering for the discriminative mixed regions leads to a lack of augmentation diversification.

\subsection{LGCOAMixer and Superpixel-attention-based Label Mixing}
\label{ccamixer}

As shown in Fig. \ref{fig2} and Algorithm \ref{algo1}, {LGCOAMixer} generates the augmented image $x_{mix}$ and its corresponding superpixel grid map $\mathbf{S_{mix}}$. The aim of superpixel-attention-based label mixing is to generate the mixed label $y_{mix}$ for the augmented image $x_{mix}$. We randomly select the number of superpixels $q_1$, $q_2$ by the uniform distribution $U(q_{min},q_{max})$ separately to achieve greater diversification. For {given} images $x_1$ and $x_2$, we obtain the corresponding superpixel maps $\mathbf{S_1}$ and $\mathbf{S_2}$ from the superpixel algorithm. Then, the superpixel-based regions (superpixels) in image $x_2$ are randomly sampled in $\mathbf{S_2}$ with Bernoulli
Distribution. We set the Bernoulli probability to $0.5$ for the largest augmentation diversification. The selected superpixels of image $x_2$ are cut and pasted into image $x_1$ for augmentation.

Similar to Eq. (\ref{eq1}), we get the augmented image $x_{mix}$ and the corresponding mixing binary mask $\mathbf{M}$, at the same time we can generate the superpixel map $\mathbf{S_{mix}}$ for the augmented image $x_{mix}$ as follows. 

\begin{equation}\label{eq4}
\begin{array}{c}
    x_{mix} = (\mathbf{1}-\mathbf{M})\odot{x_1} + \mathbf{M}\odot{x_2}\\
    \mathbf{S_{mix}} = (\mathbf{1}-\mathbf{M})\odot{\mathbf{S_1}} + \mathbf{M}\odot{\mathbf{S_2}}
\end{array}
\end{equation}
Note that when the augmented image $x_{mix}$ and the superpixel map $ \mathbf{S_{mix}}$ are generated, some superpixels in image $x_1$ may be slightly truncated because of the  possible partial overlap between the superpixels of the two images. However, 
this phenomenon is not disturbing, 
since the goal is to mix the two images.

To alleviate the issues of area-based label mixing mentioned in Sec. \ref{background}, we propose superpixel-attention-based label mixing. Instead of investigating mixing images in the image space with image concentration, we focus more on how to reduce the gap between the image space and the label space through semantic label mixing. Specifically, we generate the augmented image through random mixing of superpixels of two source images  without prejudice to diversification, and produce the augmented image label by considering the attention weights   $[w_{1}, w_2, ..., w_L]$ of all superpixel vectors (Detailed in Sec. \ref{sap} and Algorithm \ref{algo2}, Line 3). $L$ denotes the total number of superpixels in the augmented image $x_{mix}$. The semantics of each superpixel can be calculated by multiplying the attention weight by the number of pixels in the corresponding superpixel. The total semantics of the augmented image is the sum of the semantics of all superpixels. Then superpixel-attention-based $\lambda_{att}$ can be determined by the ratio of the semantics of the superpixels cut from $x_2$ and pasted into the image $x_1$ to the total semantics of the augmented image as follows.

\begin{equation}\label{eq5}
\centering
    \lambda_{att} = \frac{\sum_{i\in{\mathbf{I_{x_2}}}}w_{i}\cdot{|\mathbf{S_{mix}}[i]|}}{\sum_{j=1}^{L}w_{j}\cdot{|\mathbf{S_{mix}}}[j]|}
\end{equation}

where $\mathbf{I_{x_2}}=[I_{x_2}^1, I_{x_2}^2, ..., I_{x_2}^m]$ are the indices of superpixels cut from $x_2$ and pasted into the image $x_1$, $m$ is the number of superpixels from $x_2$ in the augmented image, $L$ is the total number of superpixels in the augmented image.

Compared to area-based label mixing \cite{yun2019cutmix, walawalkar2020attentive, kim2020puzzle}, our superpixel-attention-based label mixing method alleviates the inconsistency issue between the augmented image and the generated mixed label by mixing labels without destroying the diversification of the augmentation, using only one time of forward propagation with less computational complexity. 
We emphasize that the superpixel-attention-based mixed label, corresponding to the two classes associated with the two original images for the augmented image, is used in global classification to train a strong classifier in Sec. \ref{train-and-inference}.

\begin{algorithm}[ht!] 
\SetKwFunction{Union}{Union}\SetKwFunction{FindCompress}{FindCompress} \SetKwInOut{Input}{Input}\SetKwInOut{Output}{Output}

	\Input{Images $x_1$ and $x_2$ of size $W\times{H}$ and their corresponding label ${y_1}$ and ${y_2}$ (one-hot vectors);  the minimum and maximum  number of superpixels $q_{min}$, $q_{max}$;  superpixel selection probability $p$}
	
	\Output{Augmented image $x_{mix}$ and its mixed label $y_{mix}$ (vector of probabilities), generated superpixel map $\mathbf{S_{mix}}$
        }
\BlankLine 
	 \BlankLine 

    {$q_{1}\sim {U}(q_{min},q_{max})$, $q_{2}\sim {U}(q_{min},q_{max})$}\\
    {Superpixel map $\mathbf{S_1}$ $\leftarrow$ Superpixel algorithm($x_1$,$q_1$)}\\
    {Superpixel map $\mathbf{S_2}$ $\leftarrow$ Superpixel algorithm($x_2$,$q_2$)}\\     
    {$X\sim{B(1,p)},  P\{X=k\}=p^k(1-p)^{1-k}, k=0,1$}\\
    {$\mathbf{M}$ $\leftarrow$ $ Ind (Select(\mathbf{S_2}, X))$} \tcc{Randomly sample the superpixels in $\mathbf{S_2}$  with Bernoulli Distribution to generate a binary mask $\mathbf{M}$ }   
    {Generate $\mathbf{S_{mix}}$ and $x_{mix}$ with Eq. (\ref{eq4}) and $\mathbf{M}$} \\
    {Decoded feature $\hat{\mathbf{Z}}\in\mathbb{R}^{W\times{H}\times{D}}$ $\leftarrow$ $\theta_{dec}$($\theta_{enc}$($x_{mix}$){)}}\\
    {Superpixel weight vectors  $\mathbf{w}$ $\leftarrow$ Algorithm \ref{algo2}($\hat{\mathbf{Z}},\mathbf{S}_{mix}$)}\\ 
    {Calculate $\lambda_{att}$ with Eq. (\ref{eq5}) and $\mathbf{w}$}\\  
    {Generate $y_{mix}$ with Eq. (\ref{eq2}) and $\lambda_{att}$}
\caption{\textbf{LGCOAMixer}}
\label{algo1}
\end{algorithm}

\subsection{Superpixel Pooling, Self-Attention and Selection}
\label{sap}

\begin{figure*}[h]
    \centering	
    \subfloat[]{\includegraphics[height=3.5cm,width=5.4cm]{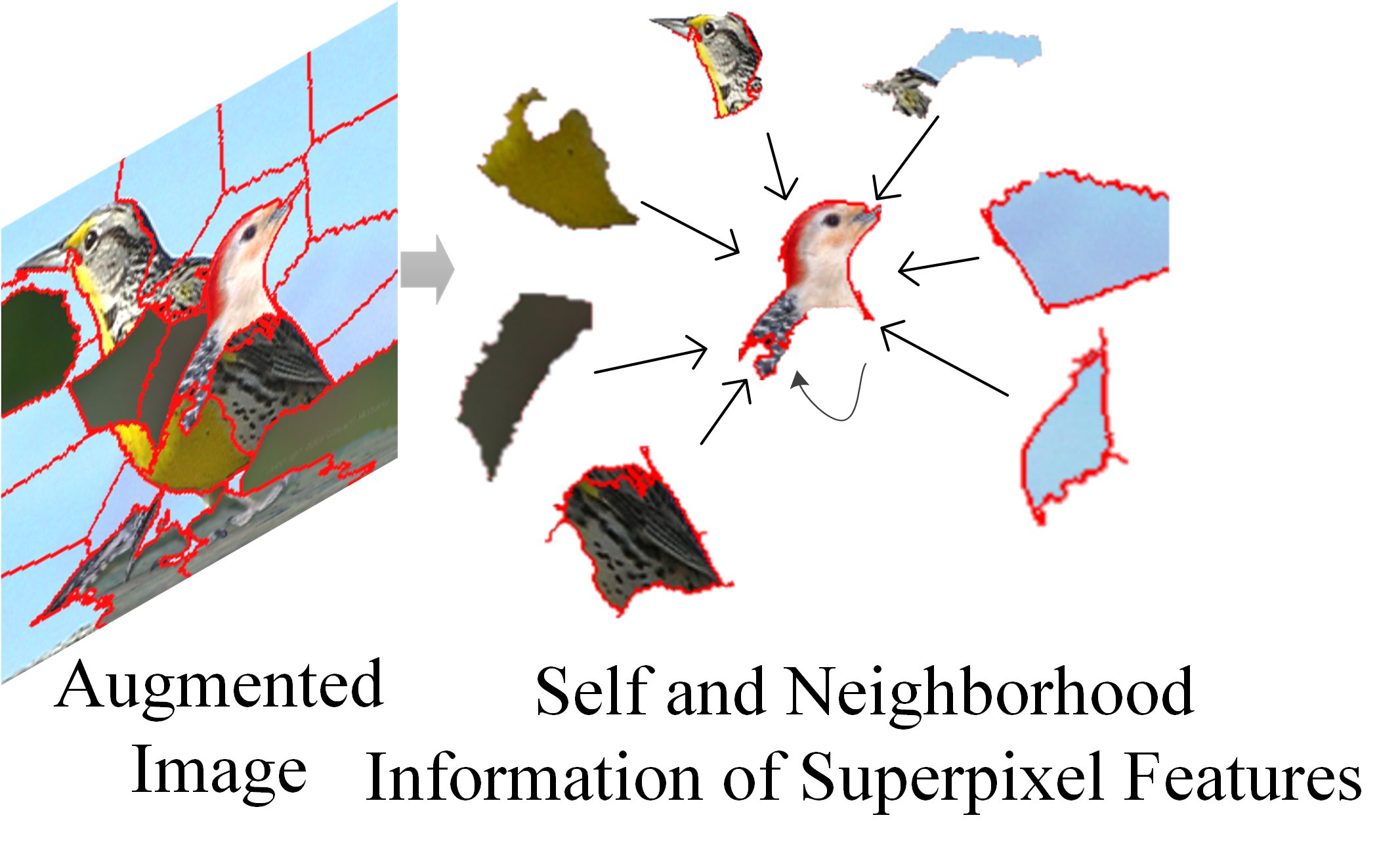}
    \label{fig3:a}}
    \subfloat[]{\includegraphics[height=4.2cm,width=8cm]{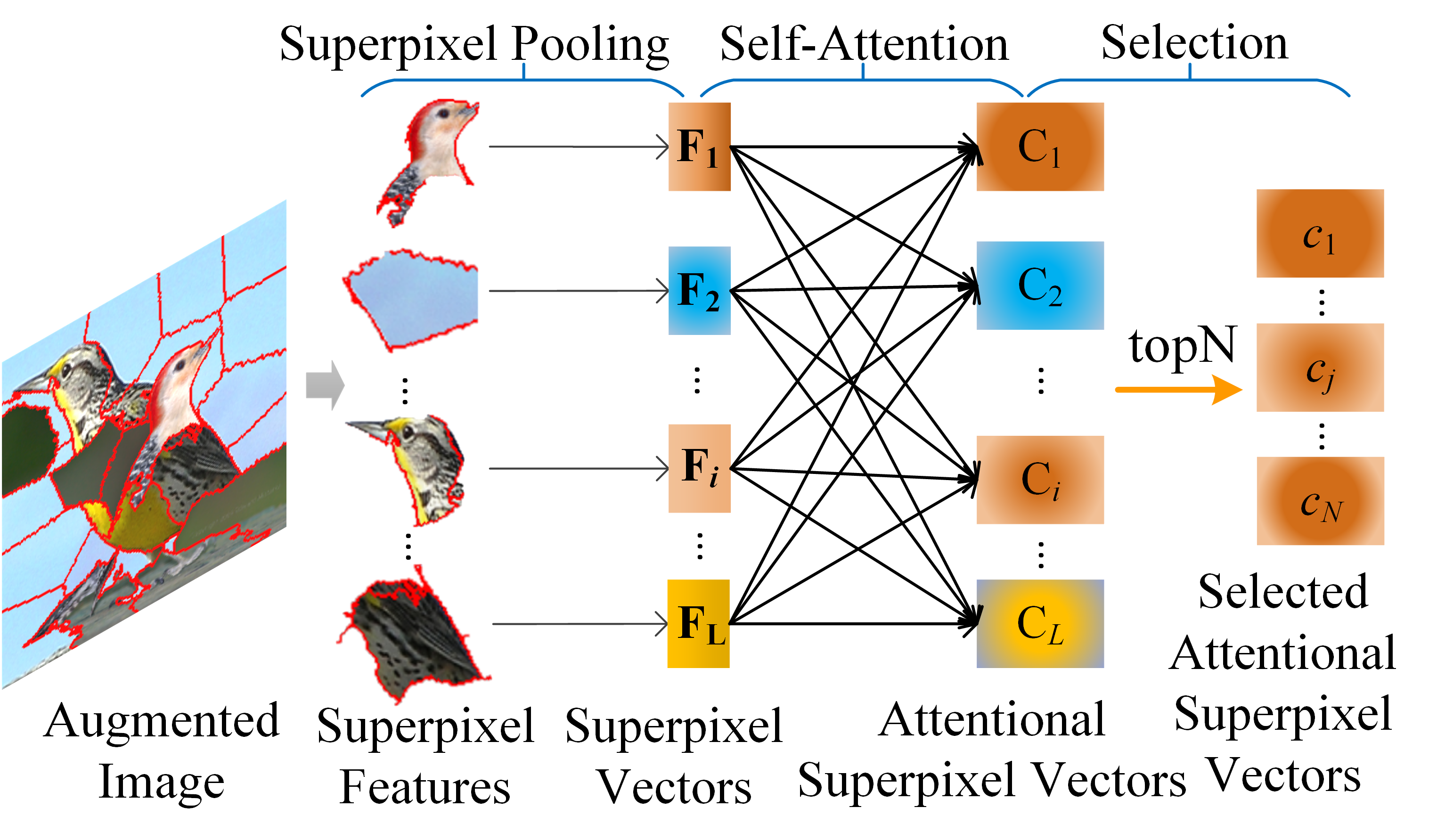}
    \label{fig3:b} }
    \caption{(a) Superpixel pooling and self-attention aims to capture local contextual and object-part information; (b) The detailed architecture of superpixel pooling and self-attention, followed by selection.}
    \label{fig3}
\end{figure*}

To preserve the object-part information in superpixel grid maps, we first decode the encoded feature $\mathbf{Z}$ to feature $\hat{\mathbf{Z}}$ with a high resolution, as shown in Fig. \ref{fig2}. To learn the object-part and contextual information of superpixels, and select the most discriminative ones, three steps are conducted, as illustrated in Fig. \ref{fig3} and Algorithm \ref{algo2}: (\romannumeral1) Superpixel pooling generating superpixel vectors, (\romannumeral2) Self-attention on superpixel vectors, and (\romannumeral3) Selection of the superpixel vectors.  

\subsubsection{Superpixel Pooling}
Superpixel pooling aims to convert the spatial footprint of the decoded feature into a sequence of superpixel feature vectors. Thus, the decoded high-resolution feature is converted into a sequence of feature vectors corresponding to the sequence of superpixels. As shown in Fig. \ref{fig3:b}, superpixel pooling performs average pooling based on the superpixel-based regions in the mixed image, rather than the square regions. In this way, superpixel pooling can preserve more object-part information compared to traditional pooling method. 

\subsubsection{Self-attention Representation}
The pooled sequenced vectors $\mathbf{F}=
[\mathbf{F}_1, \mathbf{F}_2, ..., \mathbf{F}_L]\in{\mathbb{R}^{{L}\times{D}}}$ corresponding to the sequence of superpixels are fed into a self-attention module that uses a triplet of matrices  $(\mathbf{Q, K, V})$ as follows to learn the contextual relationships.

\begin{equation}\label{eq6}
\centering
    \mathbf{Q}=\mathbf{F}\cdot{\mathbf{W}^{q}},   \mathbf{K}=\mathbf{F}\cdot{\mathbf{W}^{k}},  \mathbf{V}=\mathbf{F}\cdot{\mathbf{W}^{v}} 
\end{equation}
where $\mathbf{W}^{q}$, $\mathbf{W}^{k}$ and $\mathbf{W}^{v}\in{\mathbb{R}^{{D}\times{d}}}$ are the learnable parameters
of three linear projection matrices  and $d$ is the feature dimension in  $(\mathbf{Q, K, V})$. Self-attention (SA) is then formulated as: 

\begin{equation}\label{eq7}
\centering   
    SA(\mathbf{Q,K,V})=softmax(\frac{\mathbf{Q}\, \mathbf{K}^T}{\sqrt{d}}) \, \mathbf{V}
\end{equation}

The output sequence of feature vectors $\mathbf{C}=[\mathbf{C_1}, \mathbf{C_2}, ..., \mathbf{C_L}]\in{\mathbb{R}^{{L}\times{d}}}$ after self-attention can be described as follows. Layer normalization is applied here.

\begin{equation}\label{eq8}
\centering 
    \mathbf{C}=LayerNorm(\mathbf{F}+SA(\mathbf{Q,K,V}))
\end{equation}
In this step, the superpixel vectors are transformed into a weighted version of themselves based on the remaining superpixel feature vectors. Compared to traditional average pooling layers without attention, our superpixel pooling and self-attention module can capture local and global context information as well as object-part information, as shown in Fig. \ref{fig3:a}.

\subsubsection{Attention-based Superpixel Selection}
\label{selection}

After superpixel pooling and self-attention, we can obtain the attention weights $[w_{1}, w_2, ..., w_L]$ for all  superpixels. On the one hand, we can compute the superpixel-attention-based proportion $\lambda_{att}$ for superpixel-attention-based label mixing (described in Sec. \ref{ccamixer}, Eq. (\ref{eq5})). On the other hand, we can select the most discriminative superpixels for local superpixel classification and superpixel-wise contrastive learning. Attention-based selection allows the models to focus on the most discriminative and informative superpixels and reduce noise.

\begin{algorithm}[ht!] 
\SetKwFunction{Union}{Union}\SetKwFunction{FindCompress}{FindCompress} \SetKwInOut{Input}{input}\SetKwInOut{Output}{output}

	\Input{Decoded high resolution feature $\hat{\mathbf{Z}}\in\mathbb{R}^{W\times{H}\times{D}}$, superpixel map $\mathbf{S_{mix}}\in\mathbb{R}^{W\times{H}}$ of the augmented image, and top percentage $t$ for top superpixel selection
	}
	
	\Output{Attentional superpixel vectors $\mathbf{C}$, superpixel weight vectors $\mathbf{w}$, selected vectors $\mathbf{c_s}$ for the discriminative superpixels}
\BlankLine 

    {Vector sequence  $\mathbf{F}$ $\leftarrow$ Average pooling($\hat{\mathbf{Z}}$) by $\mathbf{S_{mix}}$ }
     
     {$\mathbf{C}\in\mathbb{R}^{L\times{d}}$ $\leftarrow $ self-attention($\mathbf{F}$)}\\ 
    {$\mathbf{w}=[w_1,w_2,...,w_L]$ $\leftarrow$ Sigmoid($\mathbf{C}.sum(dim=1)$)}\\ 

    {$\mathbf{c}_s\in\mathbb{R}^{N\times{d}}$ $\leftarrow$ top-N($\mathbf{C}$), $N=int(L\times{t})$}\\

\caption{\textbf{Superpixel Pooling, Self-Attention and Selection}}
\label{algo2}
\end{algorithm}

\subsection{Local Superpixel Classification}
\label{local-mapping}

To make the model focus more on the discriminative superpixels, we add the local superpixel classifier head, as shown in Fig. \ref{fig2}. Specifically, we perform local superpixel classification on the selected top attentional vectors of the discriminative superpixels (Described in Sec. \ref{selection}). 
And we utilize a common fully-connected layer employed by all selected superpixel vectors. This fully-connected layer utilized for local classification is distinct from the fully-connected layer utilized for global classification.
The local classification loss can be described as follows.

\begin{equation}\label{eq9}
\centering
    \mathcal{L}_{local} = \mathbb{E}\left[\sum\limits_{i=1}\limits^{N}\mathcal{H}(f_{local}(c_i),y_{s(i)}) \right]
\end{equation}

where $N$ is the number of selected superpixels in an image, $i=1,2,3,..., N$. $\mathcal{H}$ is the cross entropy loss. $c_{i}$ is the  feature vector of the  $i_{th}$ selected discriminative superpixel. $y_{s(i)}$ is the corresponding ground-truth label of the $i_{th}$ superpixel, which is either $y_1$ or $y_2$. $f_{local}$ is the fully connected layer that provides local mapping for the predicted label.

\subsection{Superpixel-wise Contrastive Learning}
\label{scl}

For improved feature embedding representation, we perform contrastive learning on the selected top attentional vectors of the discriminative superpixels in a batch. The main goal is to pull close the feature embeddings of discriminative superpixels in the same class and push apart the feature embeddings of discriminative superpixels in different classes. We achieve this by including a superpixel-wise contrastive loss \cite{khosla2020supervised} as follows.

\begin{footnotesize}  
\begin{equation}\label{eq10}
\begin{aligned}
    &\mathcal{L}_{contrast}=\\ 
    &-\frac{1}{N_B}\sum\limits_{i=1}\limits^{N_B}\frac{1}{\left|{P_i}\right|}\sum\limits_{j\in{P_i}}log\frac{exp(\mathbf{c}_i\cdot\mathbf{c}_j/\tau)}{exp(\mathbf{c}_i\cdot\mathbf{c}_j/\tau)+\sum_{k\in{N_i}}exp(\mathbf{c}_i\cdot\mathbf{c}_k/\tau)}&&
\end{aligned}
\end{equation}
\end{footnotesize}

where $\mathbf{c}_i$ denotes the unit-normalized features for superpixel $i$. $P_i$ and $N_i$ denote the corresponding positive set (superpixels in the same class) and negative set (superpixels in different classes). Note that the positive/negative superpixel features and the anchor superpixel features $i_{th}$ are not restricted to the same image, but are present throughout the whole batch, as a large number of negative examples is crucial for effective contrastive learning. $N_B$ is the number of superpixels over all images in a batch. 
We set the temperature $\tau$ as $0.7$.

\subsection{Training and Inference}
\label{train-and-inference}
Our aim is to train a strong image-level classifier, so we need essential global classification, as shown in Fig. \ref{fig2}. From the background in Sec. \ref{background}, we know the necessary label mixing in Eq. (\ref{eq2}) for the label of the augmented image. For training, when we learn from the augmented samples, the global classification loss can be described as follows.

\begin{equation}\label{eq11}
    \mathcal{L}_{global} = \mathbb{E}\left[\mathcal{H}  \left ( f_{global}(z_j), (1-\lambda_{att_j})\, y_{1_j}+\lambda_{att_j}\, {y_{2_j}} \right ) \right ]
\end{equation}
where $\mathcal{H}$ is the cross entropy loss, $z_{j}$ is the encoded features  of the  $j^{th}$ augmented image, and  $y_{1_j}$ and $y_{2_j}$ are the ground-truth labels of the source images associated with the augmented image. $\lambda_{att}$ is the computed superpixel-attention-based proportion in Eq. (\ref{eq5}) for label mixing.

For training, the objective of Eq. (\ref{eq11}) is to learn the global semantic feature of the training samples; the objective of Eq. (\ref{eq10}) is to learn a better representation of embedding space for the local superpixel-wise context with improved superpixel-wise intra-class compactness and superpixel-wise inter-class separation; the objective of Eq. (\ref{eq9}) is to strengthen the focus on the discriminative local superpixels-wise context. The three objectives are complementary to each other. Thus, the overall training loss is as follows.

\begin{equation}\label{eq12}
\centering
    \mathcal{L}_{total} = \mathcal{L}_{global} + \gamma_1 \, \mathcal{L}_{local} +\gamma_2 \, \mathcal{L}_{contrast}
\end{equation}
where $\gamma_1>0$, $\gamma_2>0$ are the two coefficients.

For inference, only the trained model for global mapping is used for classification. Therefore, we can easily infer with minimal model size.

\section{Experiments}
\label{experiments}

In this section, LGCOAMix is evaluated for classification. 
The source code is available at https://github.com/DanielaPlusPlus/LGCOAMix. 

\subsection{Datasets and Encoders}
\label{data-model}

We evaluate LGCOAMix with the datasets CIFAR100 \cite{krizhevsky2017imagenet}, TinyImageNet \cite{chrabaszcz2017downsampled}, CUB200-2011 \cite{wah2011caltech}, Stanford Dogs 
 \cite{StanfordDogs}, as shown in Table \ref{table1}. 
 It's worth noting that, the original test set from CUB200-2011 contains $5,794$ images. To load the model pre-trained on ImageNet and make the result more convincing, we remove the $23$ test images from CUB200-2011 that may overlap with the ImageNet \cite{russakovsky2015imagenet} mentioned in 
 \cite{overlapCUB}. Thus, there are $5,771$ images in the test set with $^\star$.

\begin{table}[h]
\centering
\caption{\label{table1}Standard split and input size of different datasets.}
\setlength{\tabcolsep}{10mm}{
\resizebox{0.48\textwidth}{!}{
\begin{tabular}{lrrr}
   \toprule
   \multirow{2}{*}{\textbf{Datasets}} &\multirow{2}{*}{\textbf{Input Size}} & \multicolumn{2}{c}{\textbf{Standard Split}}\\ %
   \cline{3-4} 
    &  & Training Set & Test Set \\
   \midrule
   CIFAR100&$32\times{32}$&$50,000$ & $10,000$ \\
   TinyImageNet&$64\times{64}$& $100,000$ & $10,000$ \\
   CUB200-2011&$224\times{224}$& $5,994$ & $5,771^\star$ \\
   Stanford Dogs&$224\times{224}$& $12,000$ & $8,580$ \\
\bottomrule
\end{tabular}}}
\end{table}

For the backbone encoders in the CNN structure, we utilize ResNet18, ResNet50 \cite{he2016deep}, ResNeXt50 \cite{xie2017aggregated}. We use TinyViT \cite{wu2022tinyvit} and ViT \cite{dosovitskiyimage} as transformer backbone encoders. For the CIFAR100 dataset with input size $32\times{32}$ and TinyImageNet with input size $64\times{64}$,  we modify the first convolution kernel in ResNet and ResNeXt encoders from $7\times{7}$ to $3\times{3}$.  For CUB200-2011 and Stanford Dogs, the stride of the final residual module in the ResNet and ResNeXt encoders is adjusted from $2$ to $1$. There are five transposed convolutional layers for decoding in CNN structures, three for TinyViT \cite{wu2022tinyvit}, and four for ViT \cite{dosovitskiyimage}. The encoder layers and the decoder layers are skip-connected in all CNN models (not in transformer models).

\subsection{Experimental Setup}
\label{details}

We compute the superpixel grid map for the training images with SLIC superpixel algorithm \cite{achanta2012slic}. 

On the CIFAR100 dataset, the networks are trained with a batch size of $32$ and an initial learning rate of $0.02$. The base augmentations for the training samples are random cropping for 32 with padding 4 and random horizontal flipping. For TinyImageNet, the models are trained with a batch size of $100$ and an initial learning rate of $0.02$. The base augmentation for the training samples in TinyImageNet is random horizontal flipping and random clipping for $64$ with padding $4$. For CUB200-2011, the training images are resized to $256\times{256}$ and then randomly cropped to $224\times{224}$, followed by random horizontal flipping. The test images are directly resized to $224\times{224}$. The batch size for CUB200-2011 is $8$ with an initial learning rate of $0.001$. {Training images for Stanford Dogs undergo random cropping with a ratio of $(1, 1.3)$, followed by resizing to $224\times{224}$. The images are then subject to random horizontal flipping.} The test images are directly resized to $224\times{224}$. We train the networks on Stanford Dogs with a batch size of $16$ and an initial learning rate of $0.01$.

The results obtained solely with base augmentation (cropping or flipping, etc., as mentioned above) are used as the baseline. For all other competing methods,  the corresponding augmentation scheme is used together with the base augmentation. Each image in a batch has a probability of $0.5$ to be augmented with first base augmentation and then the proposed augmentation methods, or with base augmentation only. All experiments are trained with SGD (stochastic gradient descent) optimizer with momentum as 0.9 and weight decay as $0.0005$.

\subsection{Experimental Results}
\label{results}

\begin{table}[ht!]
\caption{\label{table2} Top-1 classification accuracy on CIFAR100 using ResNet18, ResNeXt50 as encoders.}
\setlength{\tabcolsep}{1.5mm}{
\resizebox{0.49\textwidth}{!}{
    \begin{tabular}{lrcc}
    \toprule
    \multirow{2}{*}{\textbf{Method}} &
    \multirow{2}{*}{\textbf{Hyperparameters}} & \multicolumn{2}{c}{\textbf{Top-1 Acc.}}\\ 
    \cline{3-4} 
    & & \textbf{R18}  & \textbf{RX50} \\
    \midrule
     Baseline & - &  $78.58\%$ & $80.67\%$\\
    CutMix \cite{yun2019cutmix} & - & $79.69\%$ & $83.23\%$\\
    Attentive CutMix \cite{walawalkar2020attentive} & $N=3$ &$79.29\%$ & $82.51\%$\\
    SaliencyMix \cite{uddin2020saliencymix} & - & $79.57\%$ & $82.56\%$\\
    ResizeMix \cite{qin2020resizemix} & $\alpha=0.1,\beta=0.8$ & $79.71\%$ & $82.34\%$\\
    GridMix \cite{baek2021gridmix} & $grid=4\times4,p=0.8,\gamma=0.15$ &$79.45\%$ & $82.47\%$\\
    Random
    SuperpixelGridMix \cite{hammoudi2022superpixelgridmasks} & $q=200,N=50$ & $79.06\%$ & $82.22\%$\\
    Random SuperpixelGridMix \cite{hammoudi2022superpixelgridmasks} & $q=16,N=3$ & $80.30\%$ & $83.25\%$\\
    OcCaMix$\dagger$ \cite{dornaika2023object} & $q \sim U(15,50),N=3$ & $81.42\%$ & $\underline{84.01\%}$\\
    PatchUp \cite{faramarzi2022patchup}(input space) & pr=0.7,
    block =7, $\alpha=2, \gamma=0.5$ & $80.13\%$ & $83.46\%$ \\
    PatchUp \cite{faramarzi2022patchup}(hidden space) & pr=0.7,
    block =7, $\alpha=2, \gamma=0.5$ & $80.91\%$ & $83.65\%$ \\
    Saliency Grafting \cite{park2022saliency} & $\alpha=2,temperature=0.2$ & $80.83\%$ & $83.10\%$ \\
    AutoMix$\dagger$ \cite{liu2022automix} & $\alpha=2,l=3$ & $\underline{82.04\%}$ & $83.64\%$ \\
    LGCOAMix(\textbf{Ours}) & $U(30,40),0.7,\gamma_{1}=0.1, \gamma_{2}=0.05$ & $\mathbf{82.34\%}$ & $\mathbf{84.11\%}$ \\
    \bottomrule
    \vspace{-2em}
    \end{tabular}}}
\end{table}

\begin{table}[ht!]
\caption{\label{table3}Top-1 classification accuracy on TinyImageNet using ResNet18, ResNeXt50 as encoders.}
\setlength{\tabcolsep}{1.5mm}{
\resizebox{0.49\textwidth}{!}{
    \begin{tabular}{lrcc}
    \toprule
    \multirow{2}{*}{\textbf{Method}} &
    \multirow{2}{*}{\textbf{Hyperparameters}} & \multicolumn{2}{c}{\textbf{Top-1 Acc.}}\\ 
    \cline{3-4} 
    & & \textbf{R18}  & \textbf{RX50} \\ 
    \midrule
     Baseline & - &  $61.66\%$ & $65.69\%$\\
    CutMix \cite{yun2019cutmix} & - &$64.35\%$ & $66.97\%$\\
    Attentive CutMix \cite{walawalkar2020attentive} & $N=7$ & $64.01\%$ & $66.84\%$\\
    SaliencyMix \cite{uddin2020saliencymix} & - & $63.52\%$ & $66.52\%$\\
    ResizeMix \cite{qin2020resizemix} & $\alpha=0.1,\beta=0.8$ & $64.63\%$ & $67.33\%$\\
    GridMix \cite{baek2021gridmix} & $grid=8\times8,p=0.8,\gamma=0.15$ &$64.79\%$ & $67.43\%$\\
    Random
    SuperpixelGridMix \cite{hammoudi2022superpixelgridmasks} & $q=200,N=50$ & $65.59\%$ & $69.37\%$\\
    Random SuperpixelGridMix \cite{hammoudi2022superpixelgridmasks} & $q=64,N=7$ & $66.46\%$ & $71.53\%$\\
    OcCaMix$\dagger$ \cite{dornaika2023object} & $q \sim U(30,70),N=7$ & $\underline{67.35\%}$ & $\underline{72.23\%}$\\
    PatchUp \cite{faramarzi2022patchup}(input space) & pr=0.7,
    block=7, $\alpha=2, \gamma=0.5$ & $66.14\%$ & $70.49\%$ \\
    PatchUp \cite{faramarzi2022patchup}(hidden space) & pr=0.7,
    block=7, $\alpha=2, \gamma=0.5$ & $67.06\%$ & $71.51\%$ \\
    Saliency Grafting \cite{park2022saliency} & $\alpha=2,temperature=0.2$ & $64.96\%\dagger$ & $67.83\%$\\    
    AutoMix$\dagger$ \cite{liu2022automix} & $\alpha=2,l=3$ & $67.33\%$ & $70.72\%$\\
    LGCOAMix(\textbf{Ours}) & $U(25,30),0.7,\gamma_{1}=0.08,\gamma_{2}=0.05$ & $\mathbf{68.27\%}$ & $\mathbf{73.08\%}$\\
    \bottomrule
    \vspace{-2em}
\end{tabular}}}
\end{table}

\begin{table}[ht!]
\caption{\label{table4} Top-1 classification accuracy on CUB200-2011 using ResNet18, ResNeXt50 as encoders.}
\setlength{\tabcolsep}{1.5mm}{
\resizebox{0.49\textwidth}{!}{
    \begin{tabular}{lrcc}
    \toprule
    \multirow{2}{*}{\textbf{Method}} &
    \multirow{2}{*}{\textbf{Hyperparameters}} & \multicolumn{2}{c}{\textbf{Top-1 Acc.}}\\ 
    \cline{3-4} 
    & & \textbf{R18}  & \textbf{RX50} \\
    \midrule
     Baseline & - &  $75.56\%$ & $81.41\%$\\
    CutMix \cite{yun2019cutmix} & - & $76.90\%$ & $82.63\%$\\
    Attentive CutMix \cite{walawalkar2020attentive} & $N=9$ &$76.73\%$ & $82.34\%$\\
    SaliencyMix \cite{uddin2020saliencymix} & - & $76.88\%$ & $82.81\%$\\
    ResizeMix \cite{qin2020resizemix} & $\alpha=0.1$,$\beta=0.8$ & $76.23\%$ & $81.94\%$\\
    GridMix \cite{baek2021gridmix} & $14\times{14}$,$p=0.8$,$\gamma=0.15$ &$77.13\%$ & $82.17\%$\\
    Random
    SuperpixelGridMix \cite{hammoudi2022superpixelgridmasks} & $q=200$,$N=50$ & $77.58\%$ & $83.03\%$\\
    Random SuperpixelGridMix \cite{hammoudi2022superpixelgridmasks} & $q=196$,$N=9$ & $76.98\%$ & $82.19\%$\\
    OcCaMix$\dagger$ \cite{dornaika2023object} & $q\sim U(30,100)$,$N=9$ & $\underline{78.40\%}$ & $\underline{83.69\%}$\\
    PatchUp\cite{faramarzi2022patchup}(input space) & pr=0.7,
    block=7, $\alpha=2$,$\gamma=0.5$ & $77.05\%$ & $82.66\%$ \\
    PatchUp \cite{faramarzi2022patchup}(hidden space)  & pr=0.7,
    block=7, $\alpha=2$,$\gamma=0.5$ & $77.96\%$ & $83.27\%$ \\
    Saliency Grafting \cite{park2022saliency} & $\alpha=2$,$temperature=0.2$ & $77.43\%$ & $82.93\%$ \\
    AutoMix \cite{liu2022automix} & $\alpha=2$,$l=3$ & $78.17\%$ & $83.52\%$ \\
    LGCOAMix(\textbf{Ours}) & $U(30,40)$,0.7,$\gamma_{1}=0.1$,$\gamma_{2}=0.05$ & $\mathbf{78.87\%}$ & $\mathbf{84.37\%}$ \\
    \bottomrule
    \vspace{-2em}
\end{tabular}}}
\end{table}

\begin{table}[ht!]
\centering
\caption{\label{table5} Comparison of top-1 Accuracy, model size, computational complexity and inference speed on different datasets using ResNet50 as encoder.  Measured with NVIDIA GeForce RTX 2070 Super. Details on the dataset and encoder can be found  in Sec. \ref{data-model}.}
\setlength{\tabcolsep}{1.5mm}{
\resizebox{0.485\textwidth}{!}{
    \begin{tabular}{llcc|cccc}
    \toprule

    \multirow{2}{*}{\textbf{Method}}  
    & \multirow{2}{*}{\textbf{Dataset}}  & \multicolumn{2}{c}{\textbf{Training}} & \multicolumn{4}{c}{\textbf{Inference}}\\ 
    \cline{3-8} 
     & & \textbf{Param.(M)}  & \textbf{FLOPs(G)} & \textbf{Param.(M)}  & \textbf{FLOPs(G)} & \textbf{FPS}  & \textbf{Acc.} \\

    \midrule
    \multirow{3}{*}{OcCaMix\cite{dornaika2023object}}
     & CIFAR100 & 23.71 & 2.62 & 23.71 & 1.31 & 211 & $83.69\%$\\
     & TinyImageNet & 23.89 & 10.48 & 23.89 & 5.24 & 206 & $69.22\%$ \\
     & CUB200-2011 & 23.91 & 12.54 & 23.91 & 6.27 & 195 & $82.94\%$\\
    \midrule
    \multirow{3}{*}{LGCOAMix(\textbf{Ours})}
     & CIFAR100 & 37.62 & 1.95 & 23.71 & 1.31 & 211 & $83.92\%$\\
     & TinyImageNet & 37.81 & 7.80 & 23.89 & 5.24 & 206 & $70.25\%$ \\
     & CUB200-2011 & 29.71 & 9.50 & 23.91 & 6.27 & 195 & $83.56\%$\\
    \bottomrule
\end{tabular}}}
\end{table}

\begin{table}[ht!]
\caption{\label{table6} Top-1 classification accuracy on Stanford Dogs using ResNet50 as encoder.}
\setlength{\tabcolsep}{1.5mm}{
\resizebox{0.487\textwidth}{!}{
    \begin{tabular}{lrc}
    \toprule
    \textbf{Method} & {\textbf{Hyperparameters}} & \textbf{Acc.}\\ 
    \midrule
     Baseline & - &  $61.46\%$\\
    CutMix \cite{yun2019cutmix} & - &$63.92\%$\\
    Attentive CutMix \cite{walawalkar2020attentive} & $N=12$ & $62.87\%$\\
    SaliencyMix \cite{uddin2020saliencymix} & - & $64.28\%$\\
    ResizeMix \cite{qin2020resizemix} & $\alpha=0.1,\beta=0.8$ & $64.58\%$\\
    GridMix \cite{baek2021gridmix} & $14\times14,p=0.8,\gamma=0.15$ &$62.55\%$\\
    Random
    SuperpixelGridMix \cite{hammoudi2022superpixelgridmasks} & $q=200,N=50$ & $68.79\%$\\
    Random SuperpixelGridMix \cite{hammoudi2022superpixelgridmasks} & $q=196,N=12$ & $67.76\%$\\
    OcCaMix$\dagger$   \cite{dornaika2023object} & $q \sim U(50,95),N=12$ & $\underline{69.34\%}$\\
    PatchUp \cite{faramarzi2022patchup}(input space) & pr=0.7,
    block=7, $\alpha=2, \gamma=0.5$ & $64.03\%$ \\
    PatchUp \cite{faramarzi2022patchup}(hidden space)  & pr=0.7,
    block=7, $\alpha=2, \gamma=0.5$ & $65.19\%$ \\
    Saliency Grafting \cite{park2022saliency} & $\alpha=2,temperature=0.2$ & $66.32\%$\\
    AutoMix \cite{liu2022automix} & $\alpha=2,l=3$ & $69.12\%$\\
    LGCOAMix(\textbf{Ours}) & $U(40,60),0.7,\gamma_{1}=0.08,\gamma_{2}=0.04$ & $\mathbf{70.95\%}$\\
    \bottomrule
    \vspace{-1em}
\end{tabular}}}
\end{table}

\begin{table}[ht!]
\caption{\label{table7}Top-1 classification accuracy on CUB200-2011 using TinyViT-11m-224 and ViT-B/16 as encoders.}
\setlength{\tabcolsep}{1.5mm}{
\resizebox{0.487\textwidth}{!}{
    \begin{tabular}{lrcc}
    \toprule
    \multirow{2}{*}{\textbf{Method}}&
    \multirow{2}{*}{\textbf{Hyperparameters}}  & \multicolumn{2}{c}{\textbf{Top-1 Acc.}}\\ 
    \cline{3-4} \\
    & & \textbf{TinyViT11m}  & {{\textbf{ViT-B/16}}} \\
    \midrule

     Baseline & - &  $86.96\%$ & {$80.45\%$}\\
    Random
    SuperpixelGridMix \cite{hammoudi2022superpixelgridmasks} & $q=200,N=50$ & $87.19\%$ &
    {$81.32\%$}\\
    OcCaMix \cite{dornaika2023object} & $q \sim U(30,100),N=9$ & $\underline{87.88\%}$ &
    {$\underline{81.70\%}$}\\
    {LGCOAMix}(\textbf{Ours}) & $U(30,40),0.7,\gamma_{1}=0.1,\gamma_{2}=0.05$ & $\mathbf{87.99\%}$ &
    $\mathbf{82.20\%}$\\
    \bottomrule
    \vspace{-2em}
\end{tabular}}}
\end{table}

\begin{table}[ht!]
\caption{\label{table8} Top-1 classification accuracy on Stanford Dogs using TinyViT-11m-224 as encoders.}
\setlength{\tabcolsep}{1.5mm}{
\resizebox{0.487\textwidth}{!}{
    \begin{tabular}{lrc}
    \toprule
    \textbf{Method} & {\textbf{Hyperparameters}} & \textbf{Acc.}\\ 
    \midrule
     Baseline & - &  $58.92\%$\\
    Random
    SuperpixelGridMix \cite{hammoudi2022superpixelgridmasks} & $q=200,N=50$ & $59.11\%$\\
    OcCaMix \cite{dornaika2023object} & $q \sim U(50,95),N=12$ & $\underline{59.93\%}$\\
    LGCOAMix(\textbf{Ours}) & $U(40,60),0.7,\gamma_{1}=0.08,\gamma_{2}=0.04$ & $\mathbf{60.50\%}$\\
    \bottomrule
    \vspace{-3em}
\end{tabular}}}
\end{table}

The result tables also illustrate the values of the hyperparameters used. The best and second best results are shown in bold and underlined.  We tuned the Attentive CutMix \cite{walawalkar2020attentive} and Rand SuperpixelGridMix \cite{hammoudi2022superpixelgridmasks} for better performance. The hyperparameters of the other competing methods are set according to the suggestions in the corresponding paper. All the experiments on CUB200-2011 load the models pre-trained on ImageNet. The $\dagger$ marks the results which are published in the corresponding paper. {We only tuned the loss weights $\gamma_1$ and $\gamma_2$ with the 
ResNet18 encoder. Better results may be obtained by tuning the loss weights $\gamma_1$ and $\gamma_2$ with other encoder structures,  such as ResNet50, ResNeXt50, TinyViT and ViT\cite{dosovitskiyimage}.}

For all validated benchmarks, our method outperforms both classical data augmentation methods and newly published state-of-the-art methods. In Table \ref{table2}, our method outperforms the second-best scheme (i.e., AutoMix\cite{liu2022automix})  by $0.3\%$ on ResNet18, which uses a momentum pipeline to separately train the encoder and the parameterized mix block with twice forward propagation and four losses. For the TinyImage dataset results shown  in Tables \ref{table3} and  \ref{table5}, our method outperforms the second best scheme (i.e., OcCaMix\cite{dornaika2023object}) by $0.92\%$ with ResNet18, by $0.85\%$ with ResNeXt50,  and by $1.03\%$ with ResNet50. Our method also achieves the best performance on the fine-grained datasets CUB200-2011 and Stanford Dogs, both with CNN-based encoders in Tables \ref{table4} and \ref{table6}, and transformer-based encoders in Tables \ref{table7}  and \ref{table8}.

Table \ref{table5} illustrates the computational cost for the training and testing phases of the best methods: OcCaMix and our proposed method when the backbone encoder is ResNet50. It also shows the classification accuracy. It is worth noting that our method performs better than OcCaMix\cite{dornaika2023object}, which also uses superpixel grid-based mixing. This method achieves the second best performance in most cases and requires two forward propagations for both object centering and encoder training using the same encoder.

We emphasize that our approach requires the same or less time in the inference phase than all other methods, since we only use the encoder and the global classifier for inference. As can be seen in Table \ref{table5}, our method and OcCaMix\cite{dornaika2023object} have the same inference speed (FPS), although our method has more parameters (param.) for training.

\section{Ablation Study}
\label{ablation}

This section studies the effects of superpixel grid mixing (Sec. \ref{abla-superpixel}), the effects of local classification (Sec. \ref{abla-local}), the effects of superpixel contrastive learning (Sec. \ref{abla-contrastive}), the impact of the number of superpixels and top percentage for selecting discriminative superpixels (Sec. \ref{abla-nbSuperpixel}), the Bernoulli probability (Sec. \ref{abat-prob}), and compare different label mixing methods (Sec. \ref{aba-label}). We also visually compare the deep features (Sec. \ref{visual}) and evaluate our work on weakly supervised object location task (Sec. \ref{wsol}). The ablation study is summarized in Table \ref{table9}.

\begin{table}[h]
\centering
\caption{\label{table9}Ablation study of the proposed {LGCOAMix}.  {Top-1 classification Acc. on the dataset CIFAR100 was tested with ResNet18 and ResNeXt50 as the encoders. "Square." means square region-based grid mixing. "Superpixel." means superpixel region-based grid mixing. "Local-cls." means local superpixel classification. "Local-con." means local superpixel contrastive learning.}}
\setlength{\tabcolsep}{1.5mm}{
\resizebox{0.42\textwidth}{!}{
\begin{tabular}{cccc|cc}
 
    \hline
    $\textbf{Square.}$ & $\textbf{Superpixel.}$ & $\textbf{Local-cls.}$ & $\textbf{Local-con.}$ & $\textbf{R18}$ &
    $\textbf{RX50}$ \\
   \hline

   \ding{56} & \ding{56} & \ding{56} & \ding{56} & {$78.58\%$} & {$80.67\%$}\\
   
   \ding{52} & \ding{56} & \ding{56} & \ding{56} & $80.49\%$ 
   & {$83.08\%$}\\
   
   \ding{56} & \ding{52} & \ding{56} & \ding{56} & $81.27\%$ 
   & {$83.70\%$}\\
   
   \ding{56} & \ding{52} & \ding{52} & \ding{56} & $82.03\%$ 
   & {$83.94\%$}\\
   
   \ding{56} & \ding{52} & \ding{52} & \ding{52} & $\mathbf{82.34\%}$
   & {$\mathbf{84.11}\%$}\\
\hline
\vspace{-3em}
\end{tabular}}}
\end{table}

\subsection{Effects of Superpixel Grid Mixing}
\label{abla-superpixel}

Superpixel grid mixing generates augmented images based on a superpixel grid map instead of a square grid map. In Table \ref{table9} (second to third row),  the performance is improved  from $80.49\%$ to $81.27\%$ and from $83.08\%$ to $83.70\%$ compared to the square grid mixing. The reason for this improvement is that our model preserves more object-part information by utilizing superpixel grid mixing.

\subsection{Effects of Local Superpixel Classification}
\label{abla-local}

Table \ref{table9} displays the improvement achieved through the use of local superpixel classification, resulting in enhanced performance from $81.27\%$ to $82.03\%$ and from $83.70\%$ to $83.94\%$. Local superpixel loss has forced the model to extract more features from local superpixel regions.

The blue line in Fig. \ref{fig-loss-weight} shows the variation of Top 1 accuracy with the local superpixel classification loss weight $\gamma_1$. The varying of superpixel classification loss over epochs can be seen in Fig. \ref{fig-loss-total}. It can be observed that excessive focus on local regions with a high local classification loss weight can cause the model to fail in capturing global semantic information, leading to poor performance.

\subsection{Effects of Superpixel Contrastive Learning}
\label{abla-contrastive}

\begin{figure}[h!]
	\centering	
	\subfloat[]{
    \includegraphics[height=3.2cm,width=4.2cm]{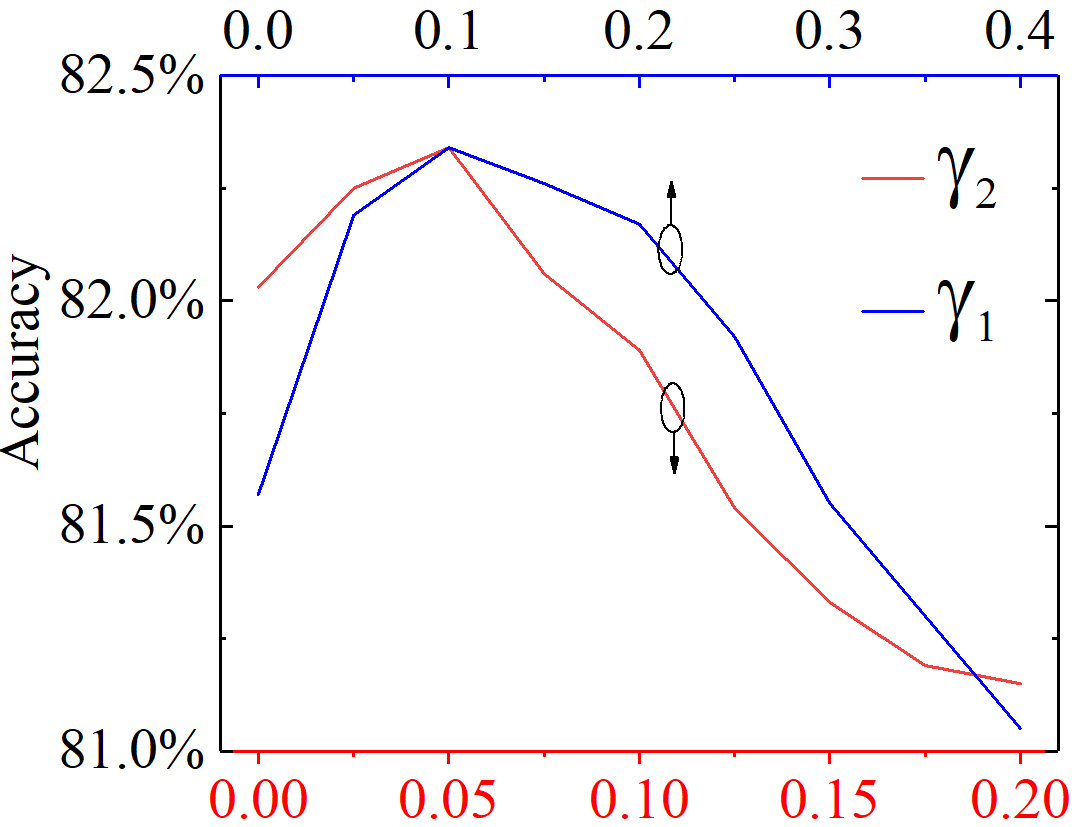}
    \centering
    \label{fig-loss-weight}}
	\subfloat[]{
    \centering
    \label{fig-loss-total} 
    \includegraphics[height=3.2cm,width=4.17cm]{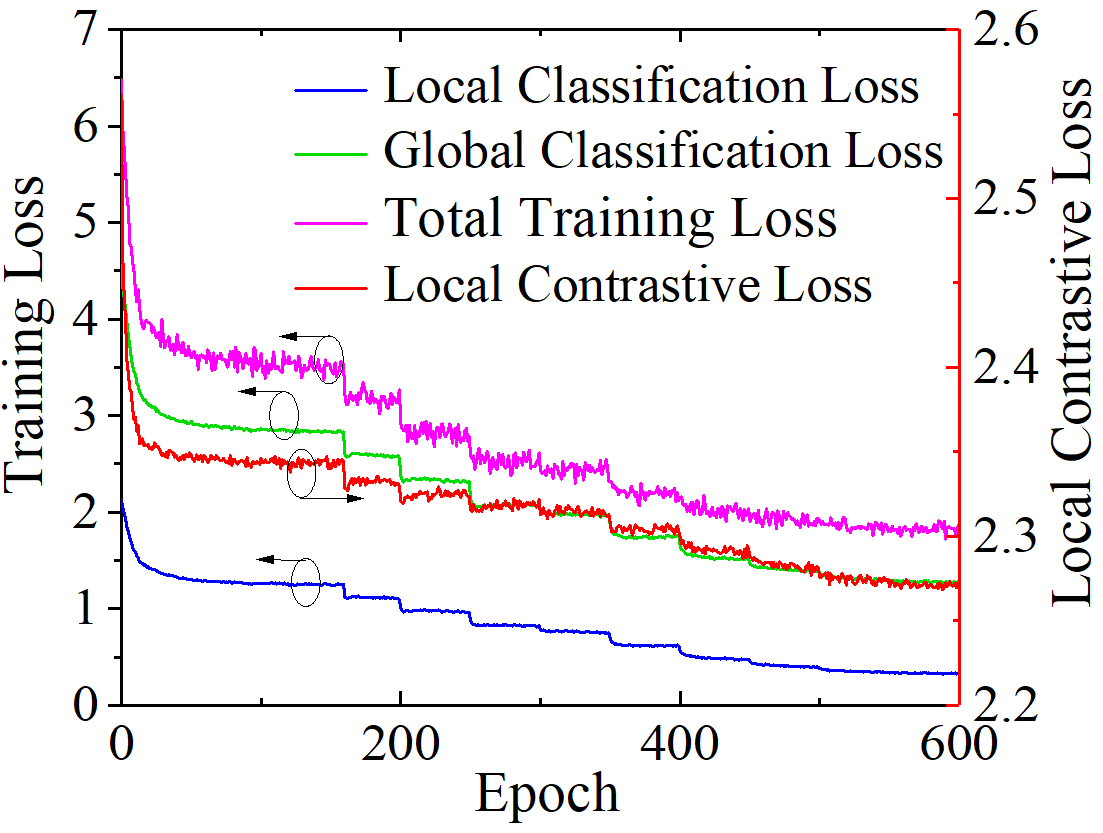}}
    \caption{{Studies of the loss weights (The quantitative performance improvements can be seen in Table \ref{table9}.) (a) Acc. for CIFAR100 with ResNet18 encoder when fixed $\gamma_{2}=0.05$ for selecting $\gamma_{1}$, and fixed $\gamma_{1}=0.1$ for selecting $\gamma_{2}$; (b) Training loss over epochs for CIFAR100 with ResNet18 encoder when fixed $\gamma_{1}=0.1$,  $\gamma_{2}=0.05$. }}\label{fig-loss}
\end{figure}

We can see that the performance is enhanced by superpixel contrastive learning, with an increase from $82.03\%$ to $82.34\%$ and from $83.94\%$ to $84.11\%$ as shown in Table \ref{table9}. The studies for the loss weight of the contrastive loss $\gamma_2$ can be found in the red line in Fig. \ref{fig-loss-weight}. The varying of superpixel contrastive loss over epochs can be seen in Fig. \ref{fig-loss-total}. It is important to note that contrastive loss is primarily intended to enhance local embeddings of superpixels. However, a high loss-weight of the contrastive loss may result in a weak classifier.

\subsection{Number of Superpixels and Top Percentage for Selecting Discriminative Superpixels}
\label{abla-nbSuperpixel}

\begin{figure}[h!]
	\centering
	\subfloat[]{\includegraphics[height=1.5cm,width=1.5cm]{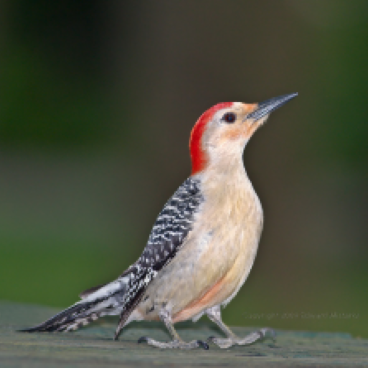}
     \label{fig4:a}  
    }
        \subfloat[]{\includegraphics[height=1.5cm,width=1.5cm]
    {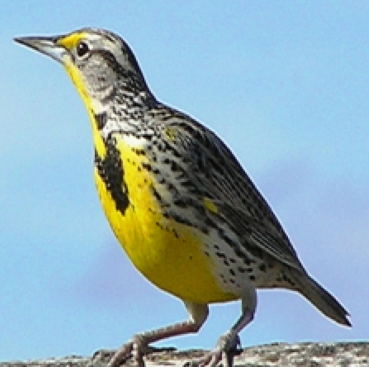}
     \label{fig4:b} 
    }
        \subfloat[$(10,10)$]{\includegraphics[height=1.5cm,width=1.5cm]{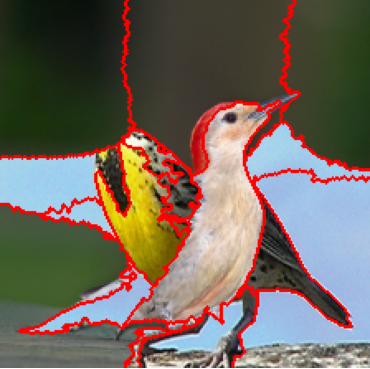}
    \label{fig4:c}  
    }
        \subfloat[$(35,35)$]{\includegraphics[height=1.5cm,width=1.5cm]{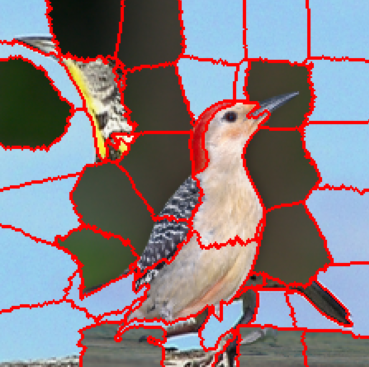}
    \label{fig4:d}  
    }
        \subfloat[$(80,80)$]{\includegraphics[height=1.5cm,width=1.5cm]{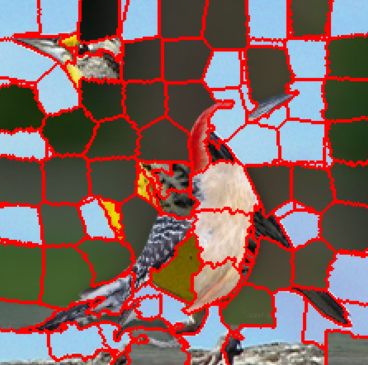}
    \label{fig4:e}  
    }
    \hfill
        \subfloat[]{\includegraphics[height=1.5cm,width=1.5cm]{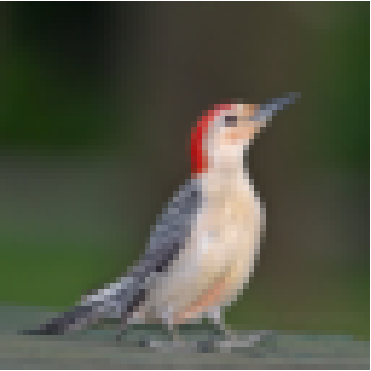}
    \label{fig4:a2} 
    } 
        \subfloat[]{\includegraphics[height=1.5cm,width=1.5cm]{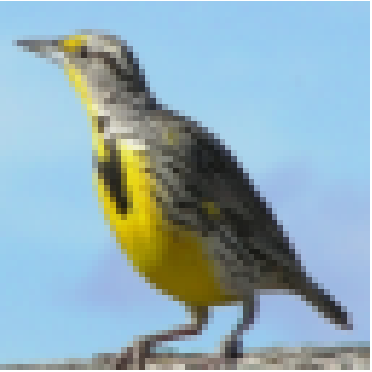}
    \label{fig4:b2}
    }
        \subfloat[$(10,10)$] {\includegraphics[height=1.5cm,width=1.5cm]{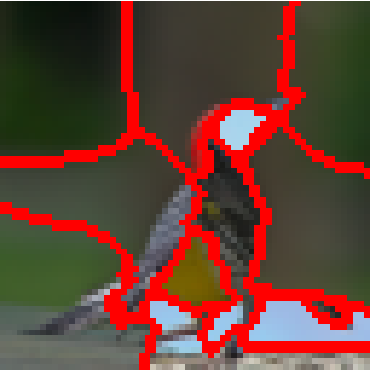}
    \label{fig4:c2} 
    }
        \subfloat[$(35,35)$] {\includegraphics[height=1.5cm,width=1.5cm]{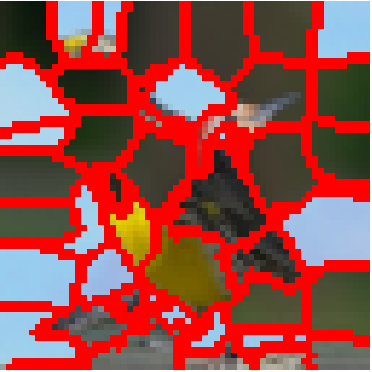}
    \label{fig4:d2}  
    }
        \subfloat[$(80,80)$] {\includegraphics[height=1.5cm,width=1.5cm]{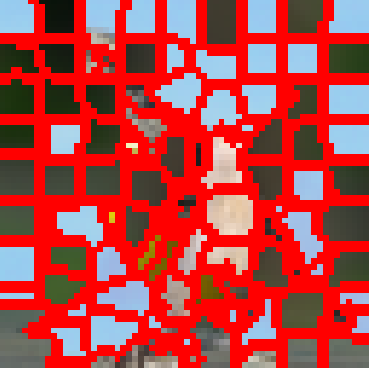}
    \label{fig4:e2}  
    }
    \caption{(a)(b) Source images of size $(224,224)$; (f)(g) Source images of size $(64,64)$; (c)-(j) are the augmented images with different input number of superpixels for source images. Note that the actual number of superpixels is not precisely equal to the input number of superpixels.}
    \label{fig4}
\end{figure}

An inadequate number of superpixels results in the loss of object-part information, while an excessive amount of superpixels results in the loss of semantic information. In our method, the number of superpixels for each image is randomly selected from a proper uniform distribution individually. In this way, we obtain the augmented samples with a larger diversification and can control the number of superpixels in a suitable range. Table \ref{table10} (upper part) shows the influence of the number of superpixels on the performance. Given a constant number of superpixels, it can be observed that some object-part information cannot be captured when the two numbers of superpixels for both source images are too small in Fig. \ref{fig4:c}. Conversely, some too-detailed object-part information without semantic meaning is captured when the two numbers of superpixels for both source images are too large in Fig. \ref{fig4:e} and Fig. \ref{fig4:e2}. Comparing Fig. \ref{fig4:c} with Fig. \ref{fig4:c2}, it is important to consider the image size when determining the number of superpixels.

\begin{table}[h]
\centering
\caption{\label{table10}Influence of number of superpixels $q$ and top percentage $t$ of the selected attention-based superpixels on top-1 classification accuracy for CIFAR100 on models with ResNet18 as the encoder.}
\setlength{\tabcolsep}{10mm}{
\resizebox{0.4\textwidth}{!}{
\begin{tabular}{llr}
   \toprule
   $q \sim U(q_{min},q_{max})$ & $t$ & \textbf{Top-1 Acc.} \\
   \midrule
   $q \sim U(10,20)$ & $70\%$ & $81.81\%$ \\
   $q \sim U(20,30)$ & $70\%$ & $82.25\%$ \\
   $q \sim U(25,30)$ & $70\%$ & $\mathbf{82.34\%}$ \\
   $q \sim U(25,35)$ & $70\%$ & $82.06\%$ \\
   $q \sim U(35,40)$ & $70\%$ & $81.77\%$ \\
  \midrule
   $q \sim U(25,30)$ & $30\%$ & $81.35\%$ \\
   $q \sim U(25,30)$ & $40\%$ & $81.43\%$ \\
   $q \sim U(25,30)$ & $50\%$ & $81.89\%$ \\
   $q \sim U(25,30)$ & $60\%$ & $82.10\%$ \\
   $q \sim U(25,30)$ & $70\%$ & $\mathbf{82.34\%}$ \\
   $q \sim U(25,30)$ & $80\%$ & $82.31\%$ \\
   $q \sim U(25,30)$ & $90\%$ & $82.15\%$ \\
\bottomrule
\end{tabular}}}
\end{table}

As described in Algorithm \ref{algo1}, we select the most discriminative local superpixel-based regions with the highest attention {weights} for local classification and contrastive learning. The top percentage for selecting is the hyperparameter $t$. Table \ref{table10} (lower part) shows the influence of the top percentage $t$ of the selected superpixels with the highest attention weight on performance. It can be seen that we obtain the best performance when the top percentage $t$ is $70\%$.

\subsection{Study the Bernoulli Probability}
\label{abat-prob}

\begin{figure}[h]
\centering
    \includegraphics[scale=0.12]{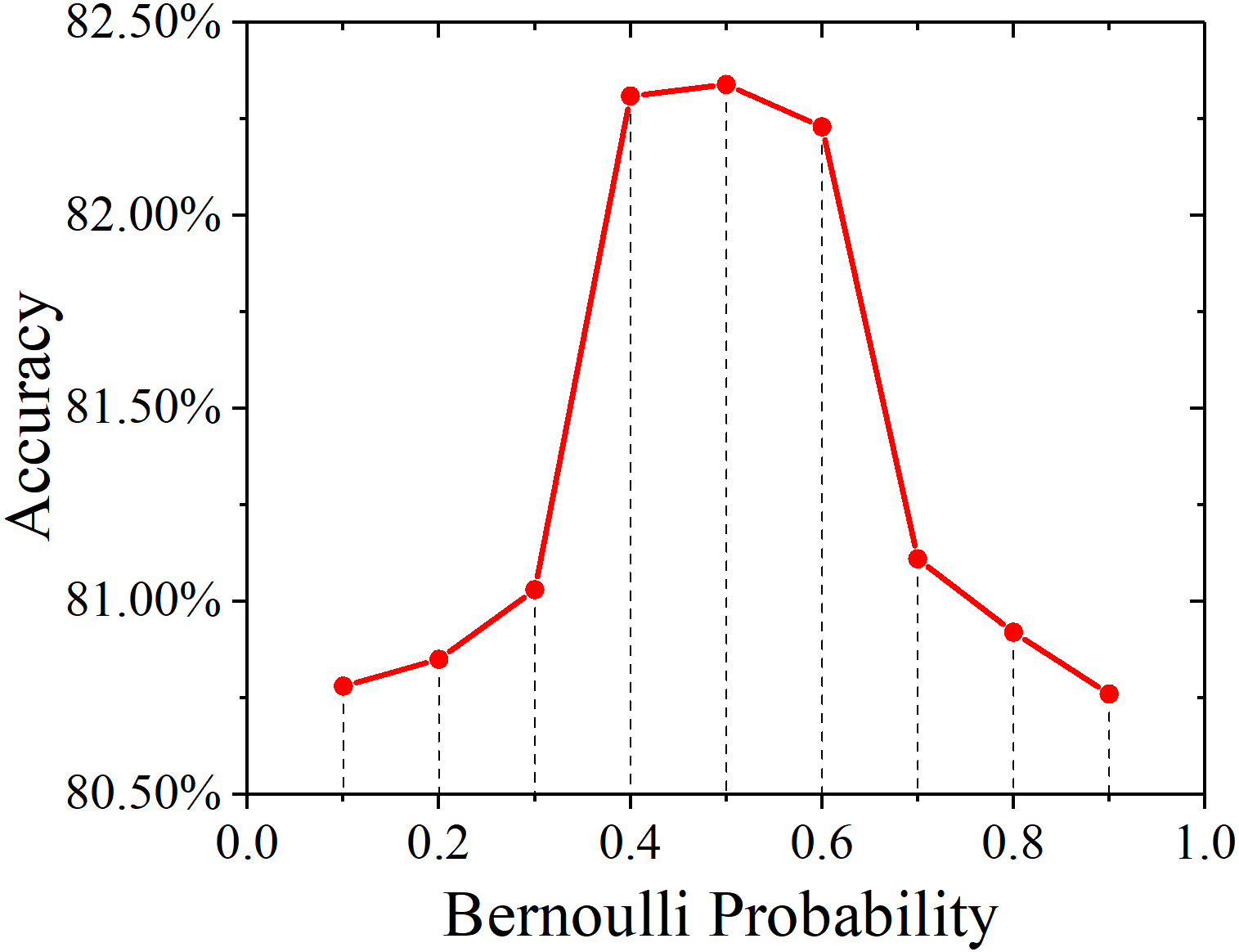}
   \caption{Top-1 classification accuracy of randomly selecting each superpixel region for mixing with different Bernoulli probabilities on CIFAR100 with ResNet18.}
   \label{fig5}
\end{figure}

During superpixel grid mixing, each superpixel region is randomly selected for mixing in the Bernoulli distribution. The randomly selected probability is the Bernoulli probability. In our procedure, we determine the randomly selected probability as $0.5$ for the most diversification of the binomial distribution on all the superpixels. Fig. \ref{fig5} also shows that the Bernoulli probability of $0.5$ can promote the performance the most.

\subsection{Comparing Different Label mixing Methods}
\label{aba-label}

There are three available label mixing methods to generate the labels for the augmented samples. In superpixel-attention-based label mixing (Ours), labels are mixed according to the attention/semantics proportion of the mixed superpixel regions; in pixel-attention-based (or saliency-based) label mixing, labels are mixed by the attention proportion of the mixed pixels; in area-based label mixing, labels are mixed by the area proportion of the mixed regions. We evaluate the performance of the above three label mixing methods in Table \ref{table11}. 
Compared to the area-based label mixing method, our superpixel-attention-based label mixing method performs better because the superpixel-attention-based label mixing method is more accurate with considering the semantic information. Compared to the pixel-attention-based label mixing method, our superpixel-attention-based label mixing method considers not only the semantic information but also the {object-part} information of the local regions.

\begin{table}[h]
\centering
\caption{\label{table11} {Top-1 classification accuracy for CIFAR100 with ResNet18 and ResNeXt50 as the encoders with different label mixing methods.}}
\setlength{\tabcolsep}{10mm}{
\resizebox{0.49\textwidth}{!}{
\begin{tabular}{lcc}
   \toprule
    \multirow{2}{*}{\textbf{Label mixing method}}  & \multicolumn{2}{c}{\textbf{Top-1 Acc.}}\\ 
    \cline{2-3} \\
    & \textbf{R18}  & {\textbf{RX50}} \\
    \midrule
   Area-based label mixing & $81.28\%$ & {$82.92\%$}\\
   Pixel-attention-based label mixing & $82.16\%$ & {$83.81\%$}\\
   Superpixel-attention-based label mixing ($\mathbf{Ours}$) & $\mathbf{82.34\%}$ & {$\textbf{84.11\%}$}\\
\bottomrule
\end{tabular}}}
\end{table}

\subsection{Visualization of Deep Features}
\label{visual}

\begin{figure}[h]
	\centering	
	\subfloat[]{\includegraphics[height=3.1cm,width=1.1cm]{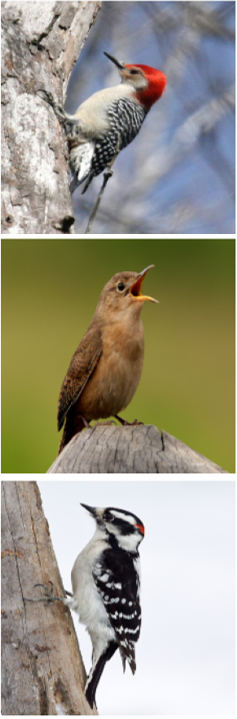}
    \label{fig-heatmap1}}    
        \subfloat[\cite{yun2019cutmix}]{\includegraphics[height=3.1cm,width=2.2cm]{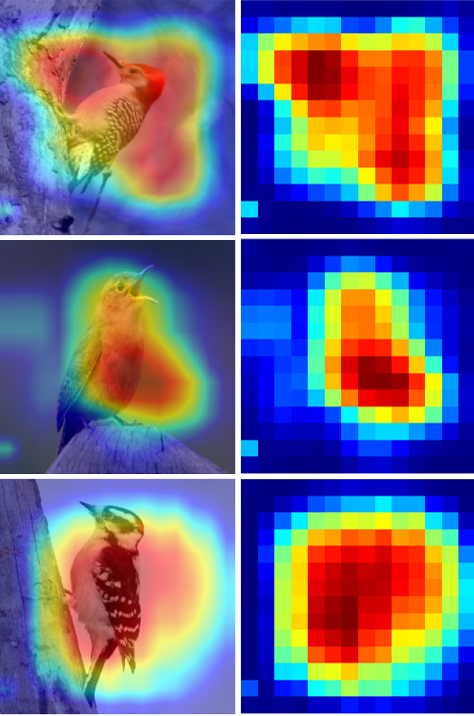}
    \label{fig-heatmap2}} 
         \subfloat[\cite{dornaika2023object}]{\includegraphics[height=3.1cm,width=2.2cm]{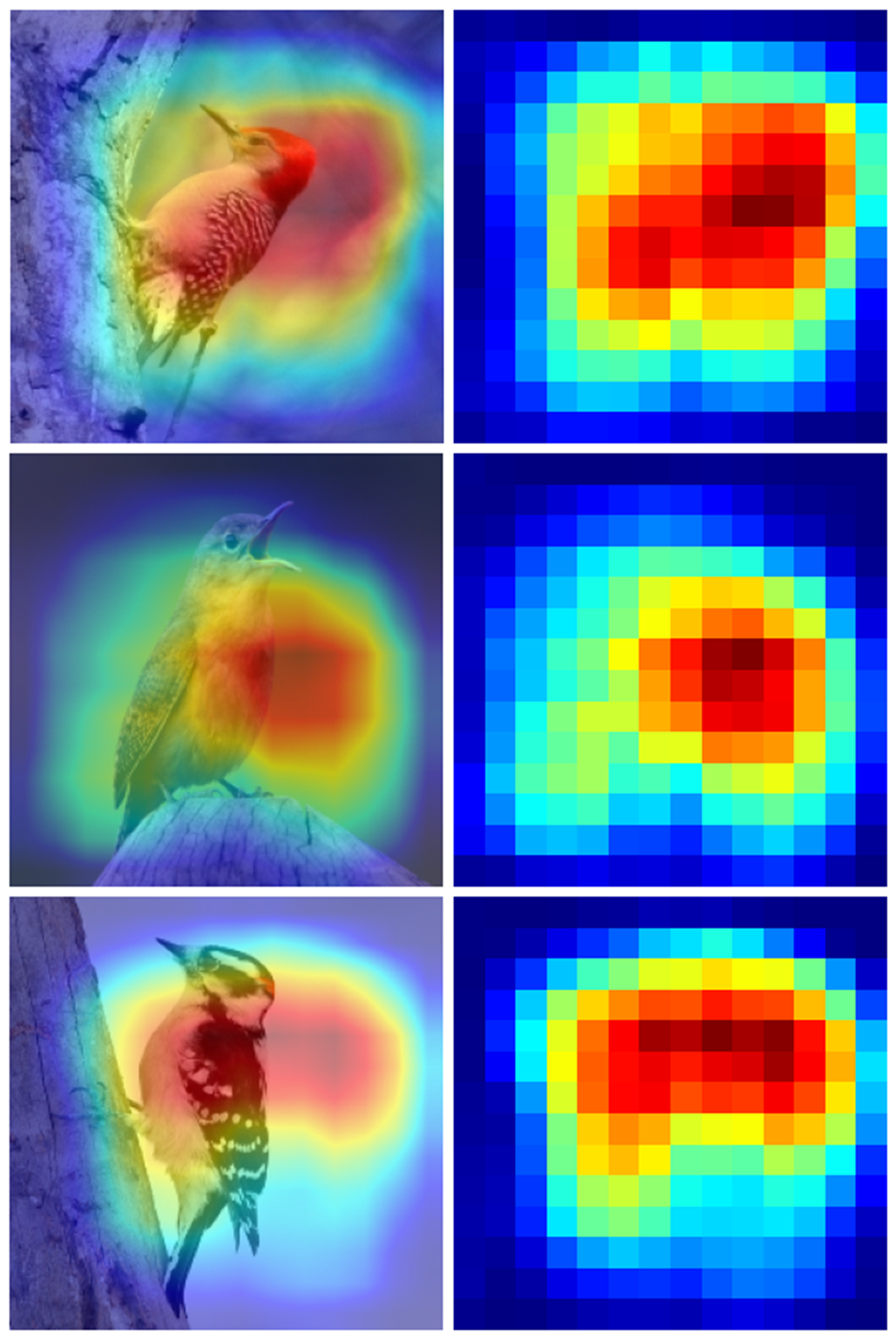}
    \label{heatmap-OcCaMix}} 
        \subfloat[Ours]{\includegraphics[height=3.1cm,width=2.2cm]{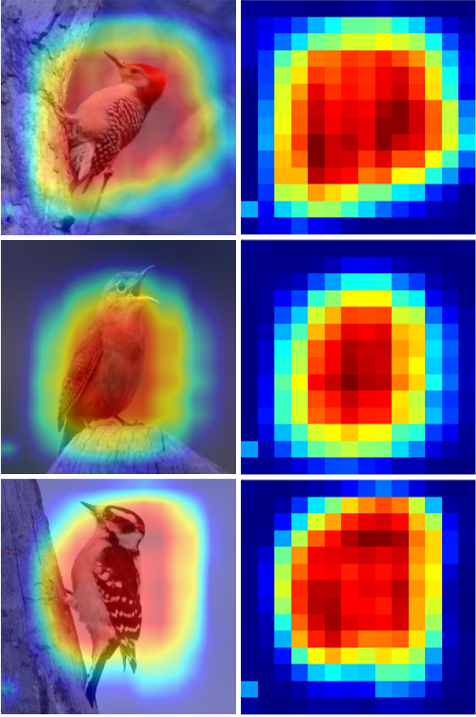}
        \label{fig-heatmap3}} 
    \caption{Visualization of the heatmaps generated by trained encoders. (a) Original images; (b)(c)(d) Heatmaps from encoder ResNeXt50 trained with CutMix\cite{yun2019cutmix}, OcCaMix \cite{dornaika2023object}, and our LGCOAMix method. All heatmaps are generated using the final encoded feature with a resolution of $14\times14$.}
    \label{fig-heatmap}
\end{figure}

\begin{figure}[h!]
	\centering
	\subfloat[Baseline]{\includegraphics[height=1.8cm,width=1.8cm]{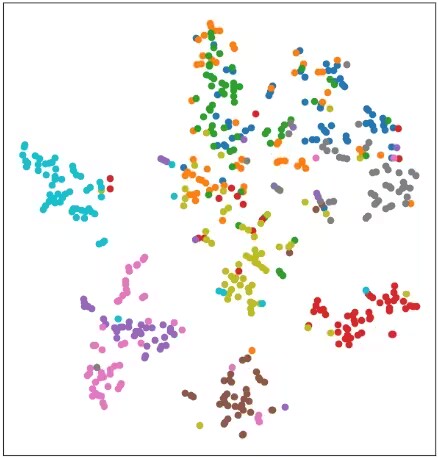}
    \label{fig-tSNE1}
    }
	\subfloat[\cite{yun2019cutmix}] {\includegraphics[height=1.8cm,width=1.8cm]{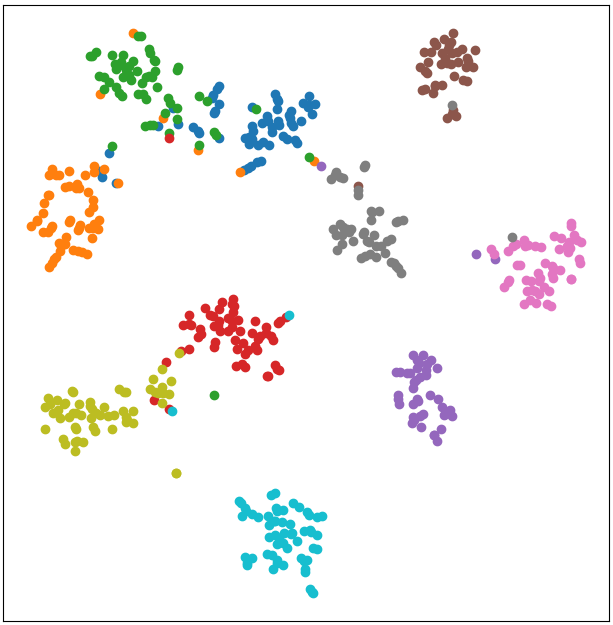}
    \label{fig-cutmix}
    }    
        \subfloat[\cite{dornaika2023object}]{\includegraphics[height=1.8cm,width=1.8cm]{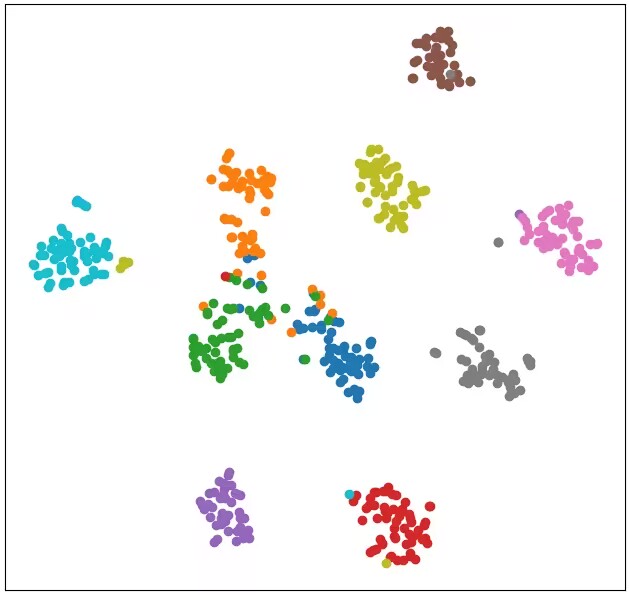}
    \label{fig-tSNE2}     
    }
        \subfloat[Ours]{\includegraphics[height=1.8cm,width=1.8cm]{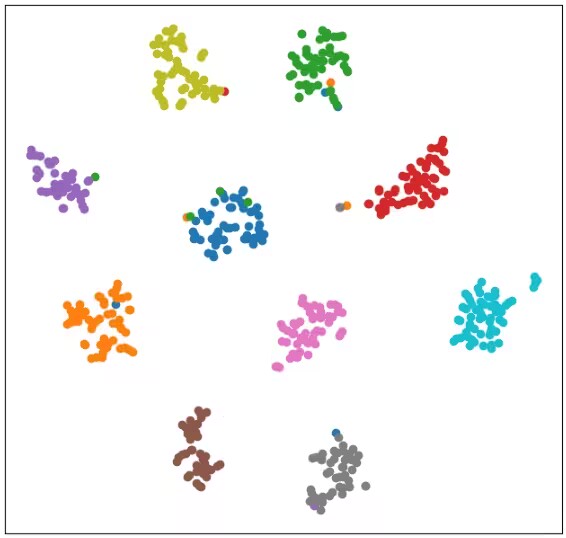}
    \label{fig-tSNE3}     
    } 
    \caption{t-SNE visualization of the CUB200-2011 features (Label 0-9). (a)(b)(c) Features extracted by ResNeXt50 trained with the baseline, CutMix \cite{yun2019cutmix}, OcCaMix\cite{dornaika2023object} and our LGCOAMix method separately.}
    \label{fig-tsne}
\end{figure}

Our proposed method is able to capture more local features and enhance the embedding of the decoded features. The results presented in Table \ref{table9} indicate that employing both local classification loss and local contrastive loss can better improve the performance of CIFAR100. Specifically, the combination of these two losses led to a boost in accuracy from $81.27\%$ to $82.34\%$ for ResNet18 and from $83.70\%$ to $84.11\%$ for ResNeXt50. Visually, Fig. \ref{fig-heatmap} shows that encoder ResNeXt50 trained with our LGCOAMix method is able to capture more local features thereby obtaining more holistic semantic information. Fig. \ref{fig-tsne} indicates that the features extracted by ResNeXt50 trained with our method achieve a better embedding.

\subsection{Weakly Supervised Object Location}
\label{wsol}

\begin{figure}[h!]
         \centering
	\subfloat[Baseline] {\includegraphics[height=3.6cm,width=2.4cm]{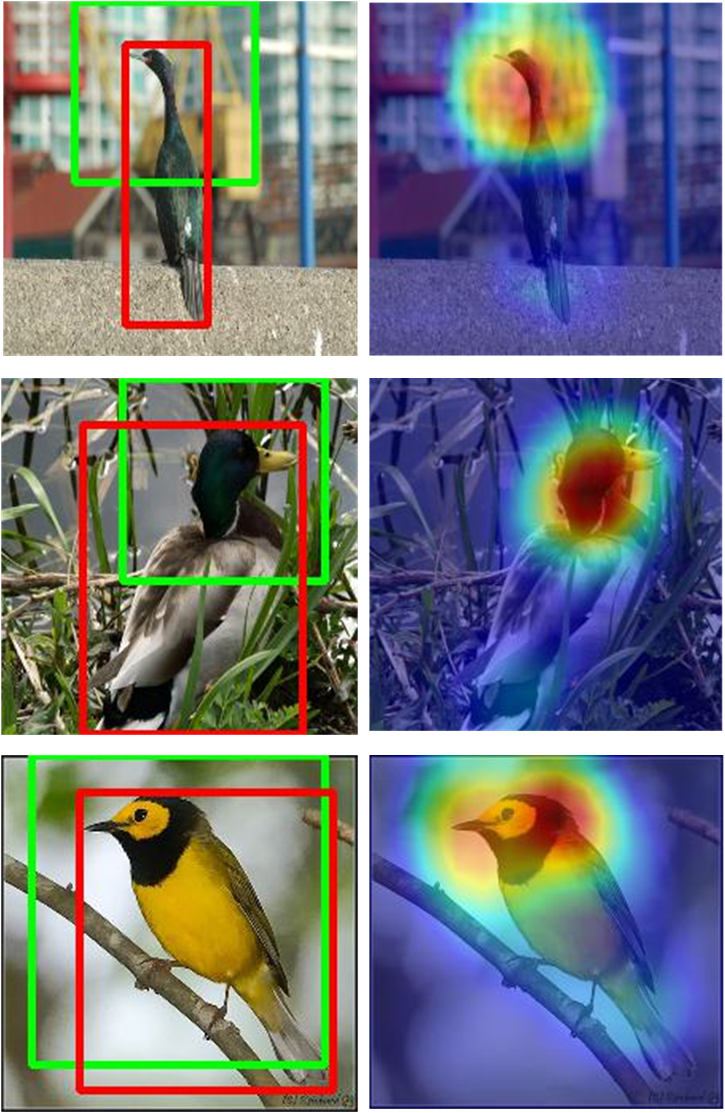}
    \label{wsol-baseline}    
    }    
        \subfloat[\cite{yun2019cutmix}]{\includegraphics[height=3.6cm,width=2.4cm]{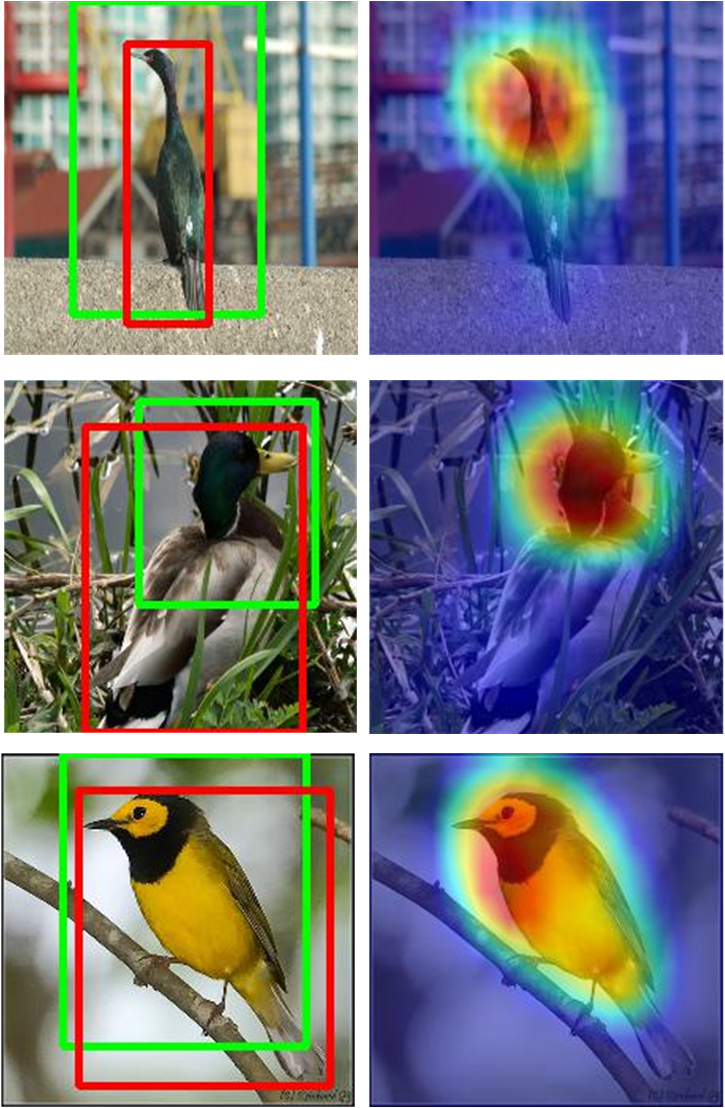}
    \label{wsol-cutmix}    
    }
        \subfloat[Ours]{\includegraphics[height=3.6cm,width=2.4cm]{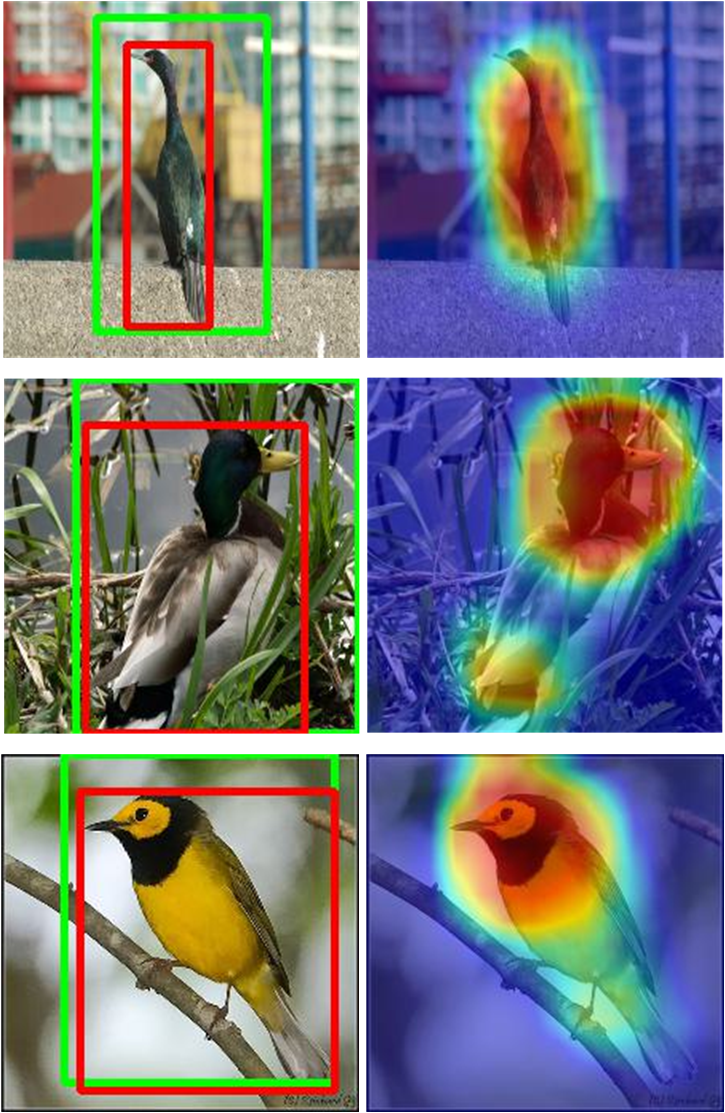}
    \label{wsol-our}
    } 
    \caption{Visulization comparison of the baseline (With no mixing based augmentation), CutMix and our method for WSOL tasks on CUB200-2011 dataset with encoder ResNet50. Ground truth and predicted bounding boxes are shown in red and green colors, respectively.}
    \label{fig-wsol}
\end{figure}

{Our proposed augmentation method and loss functions are designed for building a strong encoder for classification tasks. Nevertheless, to understand its potential for other tasks such as object detection and segmentation, we applied the proposed augmentation and the corresponding trained model to weakly supervised object location (WSOL). We followed the experimental setup described in \cite{yun2019cutmix} on CUB200-2011 and used ResNet50 as the encoder. The location accuracy defined in \cite{yun2019cutmix} is $50.21\%$, $55.22\%$, $58.65\%$ for the baseline (no mixing-based data augmentation), the CutMix method \cite{yun2019cutmix} and our proposed method. Fig. \ref{fig-wsol} shows three examples where our method can localize objects more accurately as we can capture more discriminative local features.
}

\section{Conclusion}
\label{conclusion}

In this paper, we present LGCOAMix: an efficient local and global context-and-object-part-aware superpixel-based grid mixing approach with a superpixel-attention-based semantic label blending strategy for data augmentation. We analyze the potential drawbacks of the existing cutmix-based methods for data augmentation. We also propose a novel training framework for a strong classifier that is context and object oriented. The main goal is to improve the learning of a deep encoder through image data augmentation. Extensive experiments have shown superior performance in various benchmarks and CNN models, and Transformer models. We will extend our research to semantic segmentation, {weakly supervised object location/detection}, semi-supervised learning, unsupervised learning and training for large models.


\end{document}